\documentclass{article}
\usepackage{float}
\usepackage{microtype}
\usepackage{graphicx}
\usepackage{subcaption}
\usepackage{booktabs} 

\usepackage[table]{xcolor}
\usepackage{hyperref}

\usepackage[preprint]{icml2026}

\usepackage{amsmath}
\usepackage{amssymb}
\usepackage{mathtools}
\usepackage{amsthm}

\usepackage{multirow}
\usepackage{makecell}
\usepackage{tcolorbox}

\usepackage[capitalize,noabbrev]{cleveref}

\theoremstyle{plain}

\theoremstyle{definition}

\theoremstyle{remark}

\usepackage[textsize=tiny]{todonotes}

\begin{document}

\twocolumn[
  \icmltitle{TraceVision: Trajectory-Aware Vision-Language Model for Human-Like Spatial Understanding}



  \icmlsetsymbol{equal}{*}

  \begin{icmlauthorlist}
    \icmlauthor{Fan Yang}{equal,Zidong,PCL,CASIA}
    \icmlauthor{Shurong Zheng}{equal,Zidong,PCL,CASIA}
    \icmlauthor{Hongyin Zhao}{CASIA}
    \icmlauthor{Yufei Zhan}{CASIA}
    \icmlauthor{Xin Li}{PCL}
    \icmlauthor{Yousong Zhu}{Mining}
    \icmlauthor{Chaoyang Zhao}{CASIA}
    \icmlauthor{Ming Tang}{CASIA}
    \icmlauthor{Jinqiao Wang}{Zidong,PCL,CASIA,Wuhan}
  \end{icmlauthorlist}

  \icmlaffiliation{Zidong}{Foundation Model Research Center, Institute of Automation}
  \icmlaffiliation{PCL}{Peng Cheng Laboratory, Shenzhen, China}
  \icmlaffiliation{CASIA}{School of Artificial Intelligence, University of Chinese Academy of Science, Beijing, China}
  \icmlaffiliation{Mining}{School of Artificial Intelligence, China University of Mining and Technology-Beijing, Beijing, China}
  \icmlaffiliation{Wuhan}{Wuhan AI Research, Wuhan, China}

  \icmlcorrespondingauthor{Xin Li}{xinlihitsz@gmail.com}
  \icmlcorrespondingauthor{Jinqiao Wang}{jqwang@nlpr.ia.ac.cn}

  \icmlkeywords{Machine Learning, ICML}

]



\printAffiliationsAndNotice{}  

\begin{abstract}


Recent Large Vision-Language Models (LVLMs) demonstrate remarkable capabilities in image understanding and natural language generation. However, current approaches focus predominantly on global image understanding, struggling to simulate human visual attention trajectories and explain associations between descriptions and specific regions. We propose TraceVision, a unified vision-language model integrating trajectory-aware spatial understanding in an end-to-end framework.
TraceVision employs a Trajectory-aware Visual Perception (TVP) module for bidirectional fusion of visual features and trajectory information. We design geometric simplification to extract semantic keypoints from raw trajectories and propose a three-stage training pipeline where trajectories guide description generation and region localization. We extend TraceVision to trajectory-guided segmentation and video scene understanding, enabling cross-frame tracking and temporal attention analysis.
We construct the Reasoning-based Interactive Localized Narratives (RILN) dataset to enhance logical reasoning and interpretability. Extensive experiments on trajectory-guided captioning, text-guided trajectory prediction, understanding, and segmentation demonstrate that TraceVision achieves state-of-the-art performance, establishing a foundation for intuitive spatial interaction and interpretable visual understanding.

\end{abstract}
\section{Introduction}
\label{sec:intro}

\begin{figure*}
\centering
\includegraphics[width=1\textwidth]{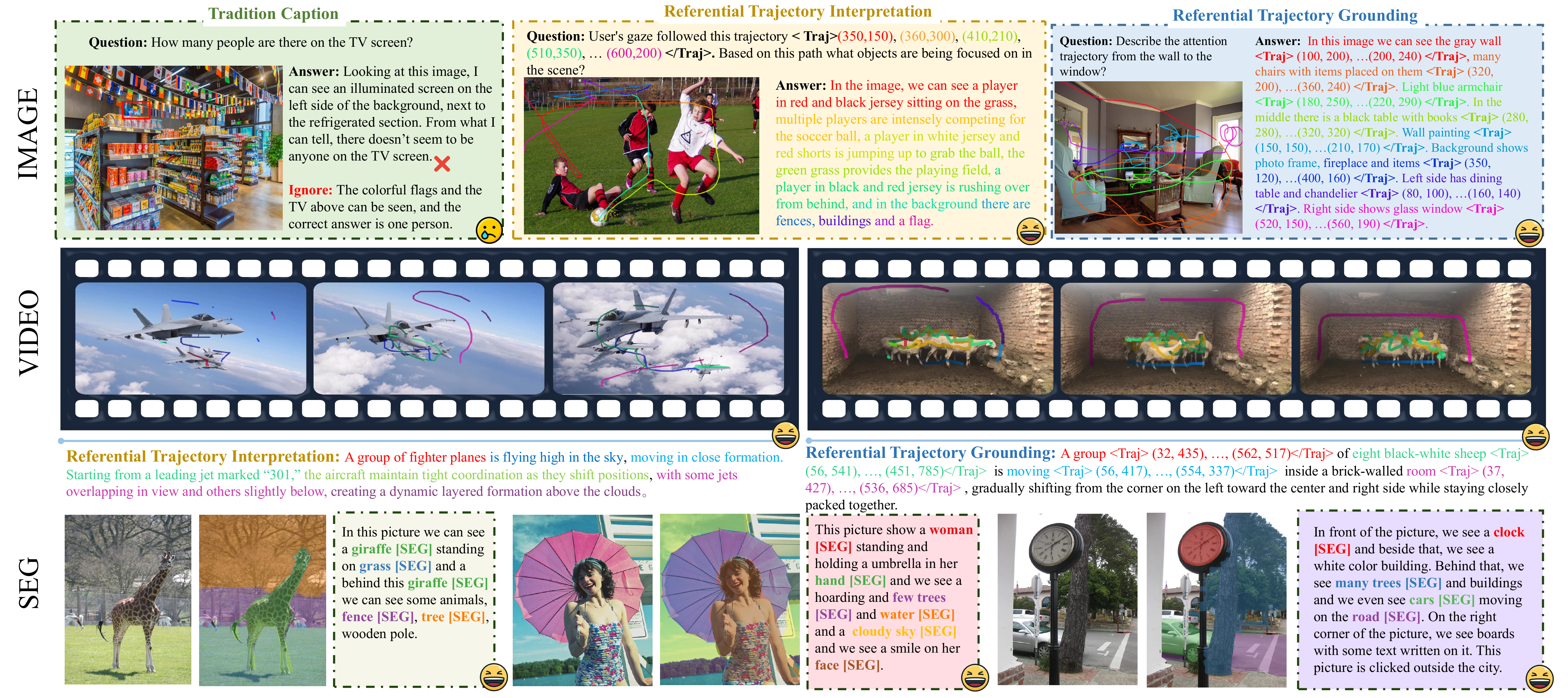}
\caption{Multi-modal capabilities of TraceVision across image, video, and segmentation tasks. The model processes traditional captioning, trajectory-guided interpretation and grounding, video sequence analysis, and precise segmentation, demonstrating its versatility in handling diverse visual understanding scenarios with trajectory-based spatial reasoning.}
\label{pre1}
\vspace{-6mm}
\end{figure*}

Emerging Large Vision-Language Models (LVLMs) \cite{llava, minigpt4, Instructblip, 2024Qwen2} make significant progress in image understanding and natural language generation, and they achieve strong performance on multimodal tasks such as Visual Question Answering (VQA) \cite{vqa, vqa2} and Image Captioning~\cite{mscoco}. However, existing LVLMs still show limitations in spatial attention modeling. As illustrated in the top-left example of Fig.~\ref{pre1}, they often focus their attention on the primary regions of an image while neglecting surrounding contextual information, and may even be distracted by irrelevant areas. In contrast, humans naturally guide their visual attention through finger movements or gestures, which helps them understand complex visual content more effectively. Research on human visual attention trajectories also plays an important role in domains such as virtual reality and autonomous driving~\cite{xia2018predicting, makrigiorgos2019human, chung2022static}.

To address this challenge, researchers have explored incorporating region information into LVLMs for controllable text generation~\cite{ferret, ferretv2, osprey, groma, psalm, controlcap}. These methods utilize regional elements such as bounding boxes, masks, points, or coordinates combined with corresponding textual prompts to guide LVLMs in generating localized captions. However, these approaches primarily rely on static discrete localization elements, struggling to model the continuity and temporal characteristics of human visual attention.  This limitation reveals several key challenges: (1) existing methods fail to capture the complex interaction between continuous spatial attention and linguistic expressions; (2) human attention trajectories involve temporal dynamics that require model understanding; (3) existing datasets lack rich trajectory-text alignment training data. This raises a fundamental research question: \textbf{\textit{how can LVLMs be enhanced to understand and respond to continuous spatial attention patterns?}}

In this paper, we present TraceVision, an end-to-end large vision-language model designed to directly predict and interpret human attention trajectories, treating them as fine-grained and temporally structured records of human attention. TraceVision models trajectories as dense and expressive signals of human intent and applies Geometric Simplification to extract semantically meaningful keypoints from raw trajectories, effectively reducing redundancy and noise while preserving geometric structures. Furthermore, we design a Trajectory-aware Visual Perception (TVP) module, which captures the sequential patterns of irregular trajectories and deeply integrates them with visual features, enabling precise localization while maintaining global context. Beyond static image tasks such as referential trajectory interpretation and referential trajectory grounding shown in Fig.~\ref{pre1}, TraceVision extends to video understanding by processing multiple frames as input to attention trajectories and to accurate region segmentation by leveraging trajectory guidance.

To ensure the robustness of TraceVision in understanding trajectory-related multi-dimensional semantic tasks, high-quality training data is essential. Existing datasets, such as Localized Narratives (LN)~\cite{LN}, provide valuable image–text–trajectory alignment but are limited to simple descriptive narratives and lack complex reasoning and instruction-following capabilities. To address this limitation, we introduce the Reasoning-based Interactive Localized Narratives (RILN) dataset, specifically designed to enhance the model logical reasoning and spatial understanding abilities. We develop an advanced data construction pipeline that leverages state-of-the-art VLMs, including GPT-4o~\cite{Openai}, Qwen2.5VL-72B~\cite{bai2025qwen2}, and Gemini-2.5 Pro~\cite{comanici2025gemini}, to automatically generate 320k high-quality instructional samples, covering diverse tasks such as referential trajectory interpretation, referential trajectory grounding, and interactive trajectory reasoning QA. Based on this dataset, we conduct a comprehensive evaluation of TraceVision. Experimental results show that TraceVision effectively exploits trajectory information, achieving state-of-the-art performance on trajectory-conditioned captioning and text-guided trajectory prediction, while also delivering competitive results on Regional Captioning and Referring Localization and Segmentation tasks.

\noindent\textbf{The main contributions of this work are:}
\vspace{0.5em}
\begin{enumerate}
\renewcommand{\labelenumi}{\textbf{\arabic{enumi}.}}
\item We propose \textbf{TraceVision}, the first end-to-end LVLM that models human attention trajectories for bidirectional trajectory–language understanding.  
\item We design a \textbf{TVP module} and a \textbf{Geometric Simplification} strategy to fuse irregular trajectories with visual features for precise spatial reasoning.  
\item We build the \textbf{RILN dataset} (320k samples) and demonstrate state-of-the-art performance on trajectory-guided captioning, prediction, and joint generation tasks.  
\end{enumerate}

\section{Related Work}
\label{sec:relate_work}



\begin{figure*}
\centering
\includegraphics[width=1\textwidth]{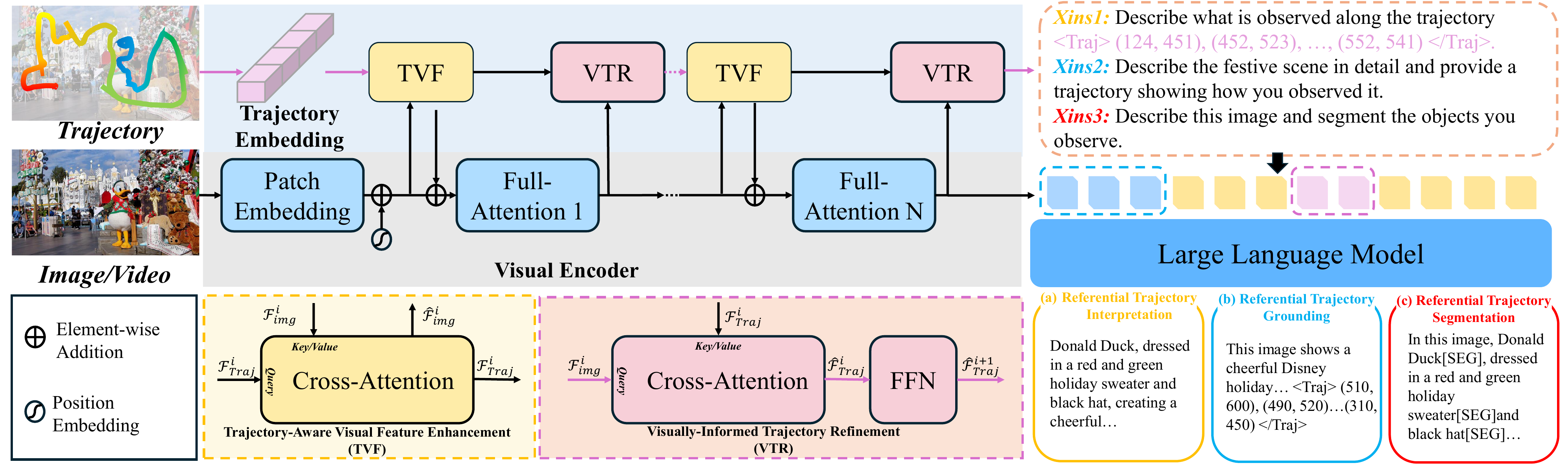}
\caption{\textbf{TraceVision architecture overview.} The model processes trajectory, image, and text inputs through a unified framework. The TVP module performs bidirectional fusion between visual and trajectory features via cross-attention, enabling trajectory-conditioned captioning, text-guided trajectory prediction tasks.}
\label{fig:main}
\end{figure*}

\vspace{-2mm}

\textbf{Vision-Language Models and Regional Understanding.} Traditional image captioning methods use generative models~\cite{chen2015mind, cornia2020meshed} with visual attention mechanisms, but these learned attentions often deviate from human perception. Controlled image captioning~\cite{xu2015show} addresses this by incorporating bounding boxes or mouse trajectories. Recent large vision-language models like Flamingo~\cite{flamingo}, BLIP2~\cite{BLIP-2}, and LLaVA~\cite{llava} excel at image-level tasks but lack precise spatial localization capabilities for region-level tasks.

\textbf{Trajectory-Guided Visual Understanding.} Visual grounding tasks focus on localizing regions based on text queries, with datasets like RefCOCO~\cite{refcoco} and Flickr30K~\cite{flickr30k}. Most methods~\cite{liu2017referring} treat this as a sparse box selection task. Localized Narratives~\cite{LN} provides dense word-region alignment through simultaneous recording of annotators' voices and mouse traces. Recent extensions include Localized Narratives Video (LNV)~\cite{LNV} for video domains and Panoptic Narrative Grounding (PNG)~\cite{PNG} for segmentation understanding.

\textbf{Human Attention Trajectory Modeling.} This emerging task aligns image descriptions with human attention trajectories collected during annotation. MITR~\cite{MITR} trains linear layers on frozen features but lacks transferability. PixelLLM \cite{0Pixel} establishes pixel-level language-vision correspondence, while \cite{xing2025large} reveals attention mechanism limitations. RegionVLM~\cite{lee2024toward} pioneers zero-shot regional understanding by converting trajectory points to text tokens, demonstrating strong performance on referring segmentation tasks. Our TraceVision proposes the first end-to-end framework for trajectory-conditioned captioning, text-guided trajectory prediction, and joint generation tasks, advancing fine-grained visual-text alignment through unified continuous spatial attention modeling.
\vspace{-2mm}

\section{Methodology}

As illustrated in Figure~\ref{fig:main}, TraceVision is a unified large vision-language model designed to process human attention trajectories bidirectionally for enhanced spatial reasoning and interpretability. Given input image $I \in \mathbb{R}^{H \times W \times 3}$, our primary objective is to establish robust correspondences between visual content, textual descriptions, and human attention patterns through trajectory-aware modeling.

\vspace{-2mm}

\subsection{Preliminary}

Traditional image captioning methods generate text sequences $S=\left[w_1, w_2, \ldots, w_n\right]$ for given images, primarily relying on global image understanding without considering spatial attention mechanisms. While existing large vision-language models perform well on image-level tasks, they remain limited in modeling human visual attention and explaining how generated descriptions correspond to specific image regions.

\textbf{Human Attention Trajectory Modeling.}
Inspired by human cognitive processes, where eye movements and gestures naturally guide visual attention, we formalize human attention trajectories as continuous temporal point sequences $T=\left\{p_1, p_2, \ldots, p_N\right\}$, with each point $p_j=(x_j, y_j, t_j)$ encoding spatial coordinates and temporal information. Unlike static, discrete localization elements, trajectories preserve the continuity and temporal dynamics of human visual exploration, providing richer signals that more faithfully reflect cognitive understanding.

\textbf{Trajectory-Text Alignment Mechanism.}
To align trajectories with language, we exploit the precise word-level temporal annotations $(w_k, t_k^{\text{start}}, t_k^{\text{end}})$ provided by the Localized Narratives dataset. For each word $w_k$, we extract its corresponding sub-trajectory $T_k$, thereby establishing dense spatial–semantic correspondences that support fine-grained multimodal understanding and bidirectional trajectory–text processing.

\textbf{Extensions to Video and Segmentation.} We further extend TraceVision to video scene understanding by introducing multi-frame visual inputs ${I_1, I_2, \ldots, I_T}$, where temporal trajectory modeling supports coherent long-sequence video description.
In segmentation tasks, we adopt a trajectory-guided strategy by introducing a special \texttt{[SEG]} token and integrating it with a segmentation decoder for prediction.
This design allows trajectories to specify interactive target regions, thereby enhancing spatial understanding and supporting fine-grained segmentation of objects of interest.

\subsection{Model Architecture}

TraceVision is built upon a unified end-to-end framework that integrates visual encoding, a large language model, trajectory processing, the Trajectory-aware Visual Perception (TVP) module, and a segmentation module to enable efficient modeling and training of human visual attention trajectories.

\vspace{-2mm}

\subsubsection{Visual Encoder} 

We employ QwenViT as the visual encoder, which effectively processes visual features at arbitrary resolutions. Given input images $I$ and a sequence of video frames $I_1, I_2, \ldots, I_t$ the encoder transforms them into comprehensive visual representations $f_{\text {visual }} \in \mathbb{R}^{N \times D}$, where $N$ denotes the number of visual tokens and $D$ represents the feature dimension. This encoding process captures rich spatial context and temporal information necessary for trajectory-visual alignment.

\vspace{-2mm}

\subsubsection{Large Language Model} 

TraceVision leverages Qwen2.5-VL-7B, a large language model that unifies the processing of visual features $f_{\text {visual }}$, textual tokens $f_{\text {text }}$ and trajectory representations $f_{\text {Traj }}$ formatted as text. This design supports trajectory-conditioned captioning, text-guided trajectory prediction and trajectory-guided segmentation.

\vspace{-2mm}

\subsubsection{Trajectory Preprocessing and Tokenization}
Raw human attention trajectories contain substantial noise and redundancy that obscure true attention patterns. We address this through a two-stage preprocessing approach combining geometric simplification and semantic tokenization.


\textbf{Semantic-Guided Geometric Simplification.} Unlike naive uniform sampling which fails to retain critical details, our approach employs a semantic-guided variant of the Douglas-Peucker (DP) algorithm that modulates sampling intensity based on the semantic weight of underlying linguistic content. We first segment the continuous trajectory $T = \left\{p_1, p_2, \ldots, p_N\right\}$ into disjoint word-aligned segments $\left\{S_1, S_2, \ldots, S_M\right\}$ guided by temporal boundaries of each word $W_i:$

\begin{equation}
S_i=\left\{p_j \in T \mid t_j \in\left[t_{\text {start }}^{(i)}, t_{\text {end }}^{(i)}\right]\right\}
\end{equation}

We employ Qwen2.5-VL-72B model to segment trajectories into semantically meaningful phrases and assign importance weights based on the model's linguistic understanding. Each phrase receives adaptive weighting that reflects its semantic significance in the overall context. For each phrase-aligned segment $S_i$, we apply the Douglas-Peucker algorithm with dynamically calculated local tolerance $\epsilon_i=\frac{\epsilon_{\text {base }}}{w_i}$, where $\epsilon_{\text {base }}$ is set to 5 pixels and $w_i$ represents normalized semantic weight of the phrase (ranging from 0.2 to 1.0, derived from importance scores of 1-5 assigned by Qwen2.5-VL). This ensures semantically important segments preserve more geometric detail critical phrases (importance 5, $w_i$=1.0) maintain maximum detail with $\epsilon_i$=5px, while minimal phrases (importance 1, $w_i$=0.2) allow aggressive simplification with $\epsilon_i$=25px.

\begin{figure}
\centering
\includegraphics[width=0.48\textwidth]{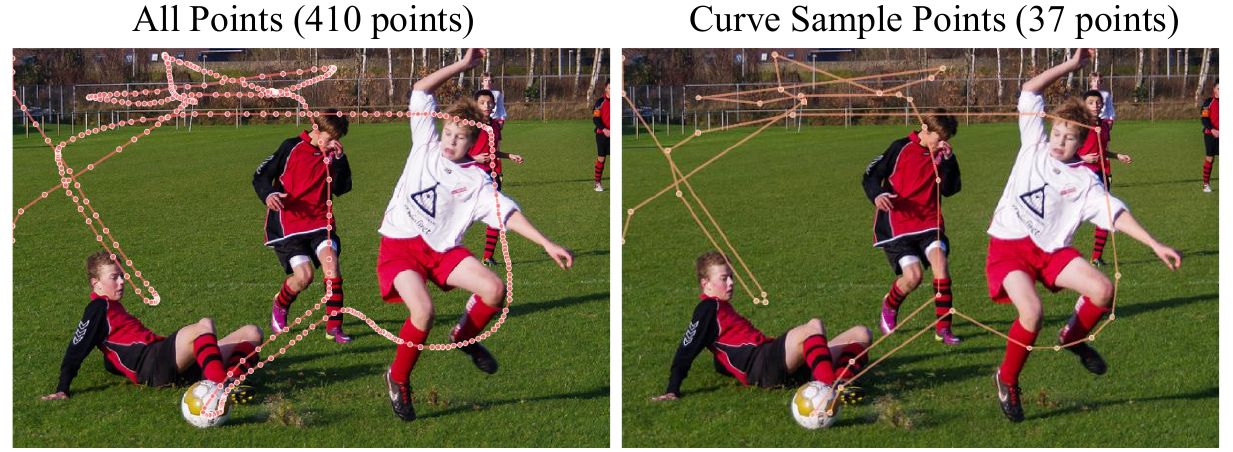}
\caption{Trajectory simplification: Geometric Simplification algorithm reduces 410 original points to 37 keypoints while preserving spatial structure.}
    \label{fig:trajectory_simplification}
\vspace{-7mm}
\end{figure}

For each trajectory segment $T_k=\left\{\left(x_i, y_i\right)\right\}_{i=s}^e$, the algorithm computes perpendicular distances from intermediate points to the line connecting endpoints:

\begin{equation}
d_i=\frac{\left|\left(y_e-y_s\right) x_i-\left(x_e-x_s\right) y_i+x_e y_s-y_e x_s\right|}{\sqrt{\left(y_e-y_s\right)^2+\left(x_e-x_s\right)^2}}
\end{equation}

Points with $d_i>\epsilon_i$ are preserved as keypoints. As is show in Fig.~\ref{fig:trajectory_simplification} this process effectively reduces trajectory density from 410 original points to 37 essential keypoints, achieving 91\% compression while maintaining spatial structure and eliminating redundant noise. The final trajectory $T^{\prime}=\bigcup_{i=1}^M \mathrm{DP}\left(S_i, \epsilon_i\right)$ maintains both geometric fidelity and semantic relevance.

\textbf{Trajectory Tokenization.} Simplified coordinates are normalized to $[0,1]$ range and quantized into 1000 discrete bins per dimension, then mapped to discrete tokens compatible with language model vocabularies. We represent coordinate pairs as formatted sequences: $\langle$ traj $\rangle\left(x_1, y_1\right),\left(x_2, y_2\right), \ldots,\left(x_n, y_n\right)\langle/$traj $\rangle$, where each coordinate value is represented as an integer token from 0 to 999, enabling seamless integration with text-based generation.

\subsubsection{Trajectory-aware Visual Perception (TVP) Module}

The TVP module performs deep bidirectional fusion of visual and trajectory information through iterative refinement. Unlike traditional methods that rely on static annotations, TVP captures temporal dynamics of human visual attention through alternating enhancement and refinement stages.

The module first tokenizes trajectory sequences $T$ and projects them into embedding space, producing trajectory features $f_{\text {Traj }}$. The core fusion process operates through two alternating stages within each block $i$:

\textbf{Trajectory-Aware Visual Enhancement.} Visual features $f_{\mathrm{img}}^i$ are enriched with trajectory guidance through cross-attention, where visual features serve as queries and trajectory embeddings as keys and values. The output is scaled by learnable parameter $\gamma^i \in \mathbb{R}^D$ (initialized to zero) for training stability:

\begin{equation}
\begin{aligned}
\hat{f}_{\mathrm{img}}^{i+1}
&= f_{\mathrm{img}}^i 
+ \gamma^i \cdot \operatorname{CrossAttn}\bigl(
Q=f_{\mathrm{img}}^i, \\
&\qquad K=f_{\mathrm{Traj}}^i,
V=f_{\mathrm{Traj}}^i
\bigr)
\end{aligned}
\end{equation}

\textbf{Visually-Informed Trajectory Refinement.} Enhanced visual features $\hat{f}_{\mathrm{img}}^i$ refine trajectory representations through a second cross-attention step, followed by feed-forward processing:


\vspace{-2mm}

\begin{equation}
\begin{aligned}
\hat{f}_{\mathrm{Traj}}^{i+1}
&= f_{\mathrm{Traj}}^i
+ \operatorname{CrossAttn}\bigl(
Q=f_{\mathrm{Traj}}^i, \\
&\qquad K=f_{\mathrm{img}}^{i+1},
V=f_{\mathrm{img}}^{i+1}
\bigr)
+ \operatorname{FFN}\bigl(
f_{\mathrm{Traj}}^i
\bigr)
\end{aligned}
\end{equation}

This bidirectional refinement creates robust multimodal embeddings that effectively integrate spatial attention patterns with visual understanding.

\begin{figure*}
    \centering
    \includegraphics[width=1\textwidth]{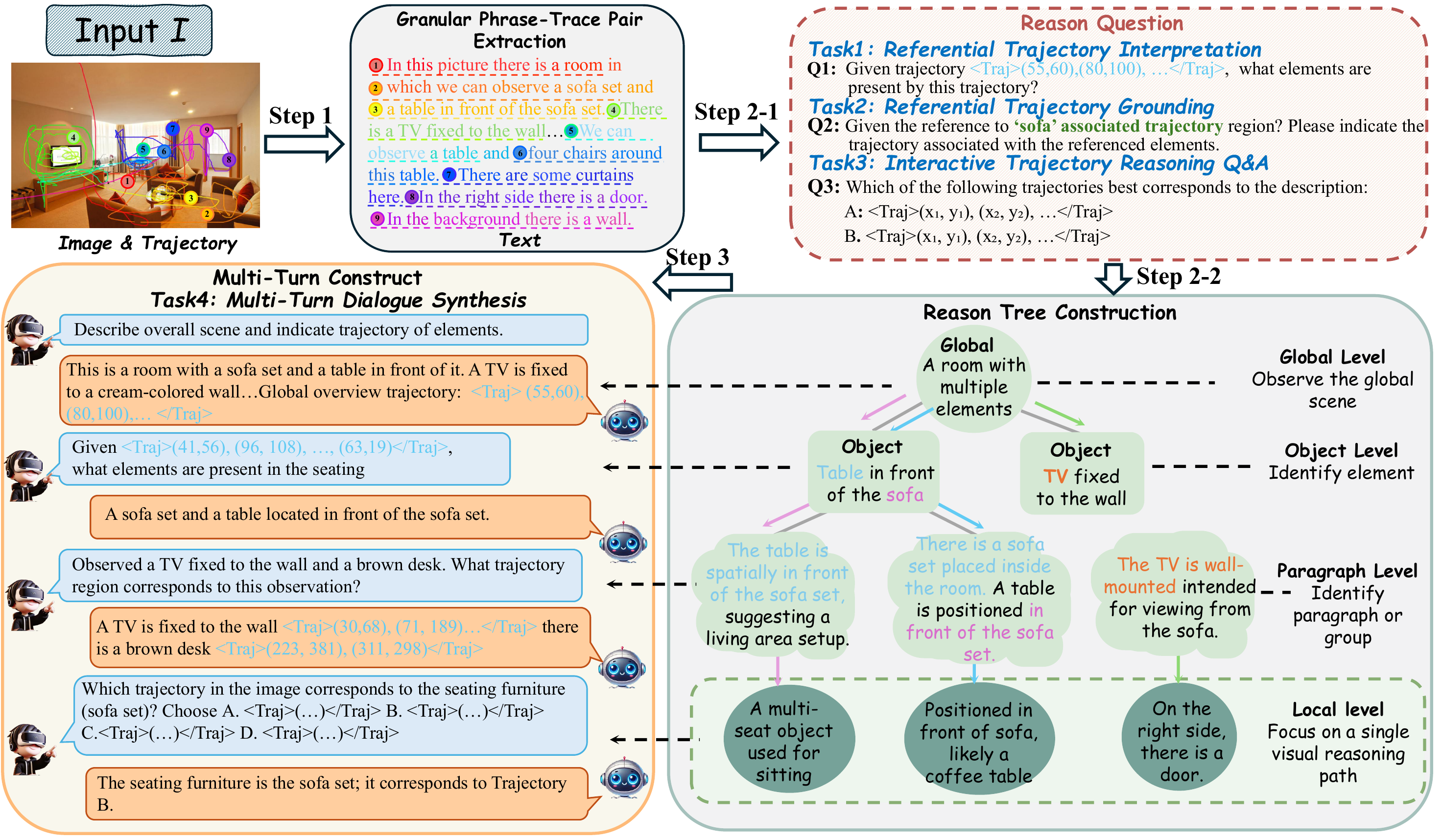}
    \caption{RILN dataset construction pipeline showing the generation of diverse trajectory-based tasks from image-trajectory pairs. The pipeline creates four main task types: referential trajectory interpretation, grounding, interactive reasoning Q\&A, and multi-turn dialogue synthesis, with hierarchical reasoning trees spanning from global scene understanding to fine-grained object-level spatial reasoning.}
    \label{fig:dataset}
    \vspace{-6mm}
\end{figure*}

\subsubsection{Segmentation Module} 
To extend TraceVision with fine-grained spatial understanding, we introduce a segmentation codebook and a lightweight segmentation decoder, following the efficient design of PixelLM.
The segmentation codebook uses 6 learnable embeddings that that encodes semantic and geometric information, which are integrated with visual features $f_{\text{visual}}$ and trajectory representations $f_{\text{Traj}}$ to provide trajectory-guided spatial priors. 
When the language model generates a \texttt{[SEG]} token, we extract its corresponding embedding as trajectory-conditioned segmentation tokens $E_{\text{seg}}$. The decoder generates pixel-level masks as:
\begin{equation}
\hat{M} = D(f_{\text{visual}}, E_{\text{seg}}),
\end{equation}

where $\hat{M}$ denotes the predicted mask and $D$ is a lightweight pixel decoder. This design enables accurate segmentation without relying on heavy decoders such as SAM and Mask2Former, and naturally generalizes to video segmentation by applying the decoder on multi-frame features $\{f_{\text{visual}}^t\}_{t=1}^T$ with trajectory guidance across frames. 



\subsection{RILN Dataset Construction}

To enhance TraceVision trajectory-aware reasoning capabilities, we develop the Reasoning-based Interactive Localized Narratives (RILN) dataset through a comprehensive data construction pipeline in Fig~\ref{fig:dataset}. While existing datasets like Localized Narratives provide basic trajectory-text alignment, they lack complex reasoning and instruction-following capabilities essential for advanced spatial understanding.

\textbf{Dataset Design and Construction Pipeline.} We build RILN based on the COCO~\cite{mscoco}, ADE20K~\cite{zhou2017scene}, Flickr30k~\cite{flickr30k}, and OpenImage~\cite{kuznetsova2020open} portions of the LN dataset, as well as video data from OVIS~\cite{qi2022occluded}, UVO~\cite{wang2021unidentifiedvideoobjectsbenchmark}, and Oops~\cite{epstein2019oops} in LNV, using only training splits to ensure evaluation independence. We employ multi-model collaborative generation to avoid single-model bias and enhance reasoning diversity, generating a total of 320k instructional samples through three stages:

\textbf{Stage 1: Visual Element and Trajectory Grounding.} We establish precise correspondences between visual elements and trajectory segments through geometric simplification for keypoint extraction, followed by word-point alignment and phrase-trajectory segmentation using Qwen2.5VL-72B.

\textbf{Stage 2: Hierarchical Reasoning Task Generation.} 
We define three core reasoning categories: referential trajectory interpretation, referential trajectory grounding, and interactive trajectory reasoning QA. 
GPT-4o generates question prompts leveraging its strength in logical coherence, while Gemini-2.5 Pro constructs structured reasoning trees and produces corresponding answers with diverse expression styles.
The reasoning tasks are organized across four cognitive levels (Global, Object, Paragraph, Local)
as a reference framework rather than rigid templates. We randomly vary expression styles, reasoning structures, and detail levels to enhance diversity and adapt to different scene complexities, ensuring a progressive transition from holistic scene understanding to fine-grained spatial analysis.

\textbf{Stage 3: Multi-Turn Dialogue Synthesis.} We synthesize 3-5 coherent dialogue sequences per image with 4-8 turns each, integrating hierarchical reasoning tasks into natural conversational flows. Final turns require synthesis of conversational context with trajectory information for comprehensive spatial understanding.

\textbf{Quality Assurance.} We employ multi-level verification, including automatic consistency checking and human expert annotation on 2000 samples, ensuring n-gram overlap (n=5) between RILN and evaluation annotations stays below 5\% to prevent textual leakage.
Models trained on RILN achieve 23\% improvement in spatial reasoning accuracy compared to baseline LN training, validating our construction methodology effectiveness.

\subsection{Training Strategy}
We adopt a three-stage curriculum learning approach that progressively builds TraceVision capabilities from basic trajectory-visual alignment to complex reasoning and dialogue understanding.

\textbf{Stage 1: Trajectory-Aware Pretraining.} We establish robust alignment between trajectory features, visual features, and language embeddings using large-scale trajectory-text-image data. Only TVP modules and trajectory embedding layers are trained, focusing on foundational cross-modal representations for trajectory-guided captioning, caption-guided trajectory prediction, and joint generation tasks.

\textbf{Stage 1.5: End-to-End Joint Training.} We unfreeze all module parameters and perform full joint training on the visual encoder, large language model, TVP module, and segmentation decoder to optimize the collaborative mechanisms between components and enhance the overall multimodal fusion capability of the system.

\textbf{Stage 2: Instruction Fine-tuning.} We employ supervised fine-tuning on RILN dataset to enable complex reasoning and conversational capabilities. Joint optimization of TVP modules, trajectory embeddings, and language decoder adapts the model to instruction-following scenarios while preserving robust multimodal representations from pretraining stages.

\section{Experiment}
\label{sec:experiment}

\subsection{Training details.} 

We build TraceVision on the Qwen2.5-VL-7B model through a three-stage training approach. In Stage 1, we freeze the visual encoder and language model while fine-tuning only the TVP module for 1 epoch, setting the batch size to 256 and learning rate to 4e-5. In Stage 1.5, we train all parameters for 3 epochs to strengthen foundational capabilities, with batch size 128 and learning rate 4e-5. In Stage 2, we freeze the visual encoder and TVP while fine-tuning only the language model using instruction data for 1 epoch, reducing the learning rate to 2e-5 and maintaining batch size at 128. We trained the model on 16 A800 GPUs for approximately 3 days total across all stages.

\begin{table*}
\caption{Quantitative results for trajectory-aware vision-language tasks on the COCO test set of Localized Narratives dataset. Controlled Caption Generation refers to generating descriptions given trajectory input, while Controlled Trajectory Generation refers to predicting trajectories given text input. Note: smaller values of LBM are better. Best results are in bold and second-best results are underlined.}
\centering
\scalebox{0.81}{
\begin{tabular}{lcccccccc}
\hline
\multicolumn{1}{l|}{}                               & \multicolumn{6}{c|}{Contrlled Caption Generation}                                                                                         & \multicolumn{2}{c}{Controlled Trace Generation} \\
\multicolumn{1}{l|}{\multirow{-2}{*}{Method}}       & BLUE-1↑         & BLUE-4↑         & METEOR↑         & ROUGE\_L↑       & CIDEr↑          & \multicolumn{1}{c|}{SPICE↑}                                  & LBM(k=0)↓                  & LBM(k=1)↓                 \\ \hline
\multicolumn{1}{l|}{LN~\cite{LN}}                             & 0.522          & 0.246          & -              & 0.483          & 1.065          & \multicolumn{1}{c|}{\underline{0.365}}                                  & -                          & -                         \\
\multicolumn{1}{l|}{MITR~\cite{MITR}}                           & 0.607          & 0.292          & 0.263          & \underline{0.487}          & 1.485          & \multicolumn{1}{c|}{0.317}                                  & 0.163                      & \underline{0.154}                     \\
\multicolumn{1}{l|}{PIxelLLM~\cite{0Pixel}}                       & -              & -              & -              & -              & -              & \multicolumn{1}{c|}{-}                                      & 0.153                      & -                         \\
\multicolumn{1}{l|}{LLaVA 1.5-13B~\cite{llava}}                      & 0.590          & 0.280          & 0.250          & 0.480          & 1.450          & \multicolumn{1}{c|}{0.310}                                  & -                      & -                     \\
\multicolumn{1}{l|}{Ferret-13B~\cite{ferret}}                      & 0.600          & 0.283          & 0.255          & 0.482          & 1.470          & \multicolumn{1}{c|}{0.315}                                  & 0.160                      & 0.170                     \\
\multicolumn{1}{l|}{Qwen2.5 VL-7B~\cite{2024Qwen2}}                     & \underline{0.630}          & \underline{0.295}          & \underline{0.260}          & 0.486          & \underline{1.500}          & \multicolumn{1}{c|}{0.320}                                  & \underline{0.147}                      & 0.168                     \\
\rowcolor{orange!30}
\multicolumn{1}{l|}{\textbf{TraceVision-7B}}  & \textbf{0.665} & \textbf{0.328} & \textbf{0.276} & \textbf{0.492} & \textbf{1.530} & \multicolumn{1}{c|}{\textbf{0.328}} & \textbf{0.117}             & \textbf{0.121}            \\ \hline

\end{tabular}}
\vspace{-3mm}
\label{tab:threetask}
\end{table*}

\begin{table*}[t]
\caption{Performance comparison on trajectory-aware regional captioning across multiple benchmarks. TraceVision demonstrates competitive performance on Visual Genome (VG), RefCOCOg, and Ref-L4 datasets, as well as benchmarks including Ferret-Bench and MDVP-Bench for spatial understanding evaluation.}
\centering
\scalebox{0.77}{
\begin{tabular}{l|cc|cc|ccc|c|c}
\hline
\multirow{2}{*}{Model} & \multicolumn{2}{c|}{VG} & \multicolumn{2}{c|}{RefCOCOg} & \multicolumn{3}{c|}{Ref-L4} & Ferret Bench & MDVP Bench \\ \cline{2-10} 
                       & METEOR      & CIDEr     & METEOR         & CIDEr        & ROUGE-L  & METEOR  & CIDEr  & Refer. Desc. & Avg.       \\ \hline
PixelLLM               &   19.9          &    \underline{148.9}       &   14.3             &   82.3           &   -       &     -    &    -    &    -          &    -        \\
GLaMM-7B~\cite{rasheed2024glamm}               & 17.0        & 127.0     & 15.7           & 104.0        & 23.8     & 10.1    & 51.1   & -            & -          \\
Osprey-7B ~\cite{osprey}             & -           & -         & 16.6           & 108.3        & -        & -       & -      & 72.2         & 44.3       \\
Ferret-7B ~\cite{ferret}             & -           & -         & -              & -            & 22.3     & 10.7    & 39.7   & 68.7         & 47.6       \\
VP-LLaVA-8B~\cite{lin2024draw}            & -           & -         & 22.4           & 153.6        & -        & -       & -      & 75.2         & 70.6       \\
VP-SPHINX-7B~\cite{lin2024draw}          & 20.1        & 139.5     & 21.0           & 138.7        & 22.5     & 10.5    & 32.2   & 73.1         & 71.5       \\
VP-SPHINX-13B~\cite{lin2024draw}          & 20.6        & 141.8     & 23.9           & \underline{162.5}        & 22.6     & 10.7    & 32.4   & \underline{77.4}         & \textbf{74.3}       \\
Omni-RGPT-7B~\cite{heo2025omni}           & 17.0        & 139.3     & 17.0           & 109.7        & -        & -       & -      & -            & -          \\
RegionGPT-7B~\cite{guo2024regiongpt}           & 17.0        & 145.6     & 16.9           & 109.9        & 25.3     & 12.2    & 42.0   & -            & -          \\
DAM-8B~\cite{lian2025describe}                 & -           & -         & -              & -            & \textbf{37.1}     & \underline{19.4}    & \textbf{70.0}   & -            & -          \\
PAM-3B~\cite{lin2025perceive}                 & \underline{20.8}        & 142.3     & \underline{26.9}           & 143.1        & 31.3     & 17.2    & 59.7   & \textbf{77.5}         & 72.2       \\
\rowcolor{orange!30}
\textbf{TraceVision-7B}            &    \textbf{21.5}         &   \textbf{163.2}        &   \textbf{28.8}             &    \textbf{168.2}          &   \underline{36.5}       &   \textbf{20.3}      & \underline{69.7}       & 76.7             &    \underline{73.1}        \\ \hline
\end{tabular}}
\label{region}
\vspace{-2mm}
\end{table*}

\begin{table}[t]
\captionof{table}{Performance comparison on referring localization and segmentation across RefCOCO datasets. TraceVision achieves state-of-the-art performance on both bounding box localization (P@0.5) and segmentation (cIoU) tasks. $^*$ indicates zero-shot results without task-specific fine-tuning.}
\centering
\scalebox{0.58}{
\begin{tabular}{lcccccccc}
\hline
\multicolumn{1}{l|}{\multirow{2}{*}{Models}} & \multicolumn{3}{c|}{RefCOCO}                                       & \multicolumn{3}{c|}{RefCOCO+}             & \multicolumn{2}{c}{RefCOCOg} \\
\multicolumn{1}{l|}{}                        & val           & testA         & \multicolumn{1}{c|}{testB}         & val  & testA & \multicolumn{1}{c|}{testB} & val           & test         \\ \hline
\multicolumn{9}{c}{bounding box P @ 0.5}                                                                                                                                                     \\ \hline
\multicolumn{1}{l|}{VisionLLM~\cite{wang2023visionllm} }              & 86.7          & -             & \multicolumn{1}{c|}{-}             & -    & -     & \multicolumn{1}{c|}{-}     & -             & -            \\
\multicolumn{1}{l|}{Shikra-7B~\cite{chen2023shikra}}               & 87.0          & 90.6          & \multicolumn{1}{c|}{80.2}          & 81.6 & 87.4  & \multicolumn{1}{c|}{71.1}  & 82.3          & 82.2         \\
\multicolumn{1}{l|}{Ferret-7B}               & 87.5          & 91.4          & \multicolumn{1}{c|}{82.5}          & 80.8 & 87.4  & \multicolumn{1}{c|}{73.1}  & 83.9          & 84.8         \\
\multicolumn{1}{l|}{PixelLLM-7B}                & 89.8          & 92.2          & \multicolumn{1}{c|}{\underline{86.4}}          & 83.2 & 87.0  & \multicolumn{1}{c|}{\underline{78.9}}  & 84.6          & 86.0         \\
\multicolumn{1}{l|}{Qwen2.5-VL-7B}              & \underline{90.0}          & \underline{92.5}          & \multicolumn{1}{c|}{85.4}          & \underline{84.2} & \textbf{89.1}  & \multicolumn{1}{c|}{76.9}  & \textbf{87.2}          & \underline{87.2}         \\
\rowcolor{orange!30}
\multicolumn{1}{l|}{TraceVision-7B}                & \textbf{90.4} & \textbf{93.1} & \multicolumn{1}{c|}{\textbf{87.8}} & \textbf{84.6} & \underline{88.2} & \multicolumn{1}{c|}{\textbf{80.5}} & \underline{86.9} & \textbf{88.3} \\ \hline
\multicolumn{9}{c}{segmentation mask cIoU}          \\ \hline
\multicolumn{1}{l|}{RegionVLM-4B$^*$~\cite{lee2024toward}} & 38.7           & 39.4          & \multicolumn{1}{c|}{37.6}          & 31.5 & 34.0  & \multicolumn{1}{c|}{30.2}  & 33.9         & -         \\
\multicolumn{1}{l|}{LISA-7B~\cite{lai2024lisa}}                    & 74.9          & 79.1          & \multicolumn{1}{c|}{72.3}          & 65.1 & 70.8  & \multicolumn{1}{c|}{58.1}  & 67.9          & 70.6         \\
\multicolumn{1}{l|}{PixelLLM-7B}                & 76.9          & 78.5          & \multicolumn{1}{c|}{74.4}          & 69.2 & 72.1  & \multicolumn{1}{c|}{64.5}  & 70.7          & 72.4         \\
\multicolumn{1}{l|}{PixelLM-7B~\cite{ren2024pixellm}}              & 73.0          & 76.5          & \multicolumn{1}{c|}{68.2}          & 66.3 & 71.7  & \multicolumn{1}{c|}{58.3}  & 69.3          & 70.5         \\
\multicolumn{1}{l|}{PSALM-3B~\cite{psalm}}                & \underline{83.6}          & 84.7          & \multicolumn{1}{c|}{81.6}          & 72.9 & 75.5  & \multicolumn{1}{c|}{70.1}  & 73.8          & 74.4         \\
\multicolumn{1}{l|}{HyperSeg-1.5B~\cite{nirkin2021hyperseg}}           & \textbf{84.8}          & \underline{85.7}          & \multicolumn{1}{c|}{\textbf{83.4}}          & \underline{79.0} & \underline{83.5}  & \multicolumn{1}{c|}{\underline{75.2}}  & \textbf{79.4}          & \underline{78.9}         \\
\rowcolor{orange!30}
\multicolumn{1}{l|}{TraceVision-7B}             & 83.4 & \textbf{86.8} & \multicolumn{1}{c|}{\underline{82.4}} & \textbf{80.1} & \textbf{84.2} & \multicolumn{1}{c|}{\textbf{76.8}} & \underline{77.6} & \textbf{80.1} \\ \hline
\end{tabular}}
\label{tab:results_comparison}
\vspace{-7mm}
\end{table}

\subsection{Comparison Controlled Caption and Trace generation}

In the table ~\ref{tab:threetask}, we evaluate two trajectory-aware tasks: Controlled Caption Generation (generating descriptions from images and trajectories) and Controlled Trajectory Generation (predicting trajectories from text descriptions).

\textbf{Controlled Caption Generation.} Given an image and trajectory, the model generates corresponding captions evaluated using BLEU, METEOR, ROUGE-L, CIDEr, and SPICE metrics. 
We compare against LVLMs using task-specific prompts and PixelLLM as baseline. 
TraceVision consistently outperforms all methods, demonstrating superior trajectory-conditioned description quality.

\textbf{Controlled Trajectory Generation.} The model predicts continuous point sequences given input captions, evaluated using LBM scores (k=0, k=1) for spatial alignment accuracy. We compare against MITR (bounding box sequences) and adapted LVLMs with prompted trajectory generation. TraceVision outperforms baseline methods in LBM scores, validating its trajectory prediction effectiveness.

\begin{figure}
    \centering
    \includegraphics[width=0.48\textwidth]{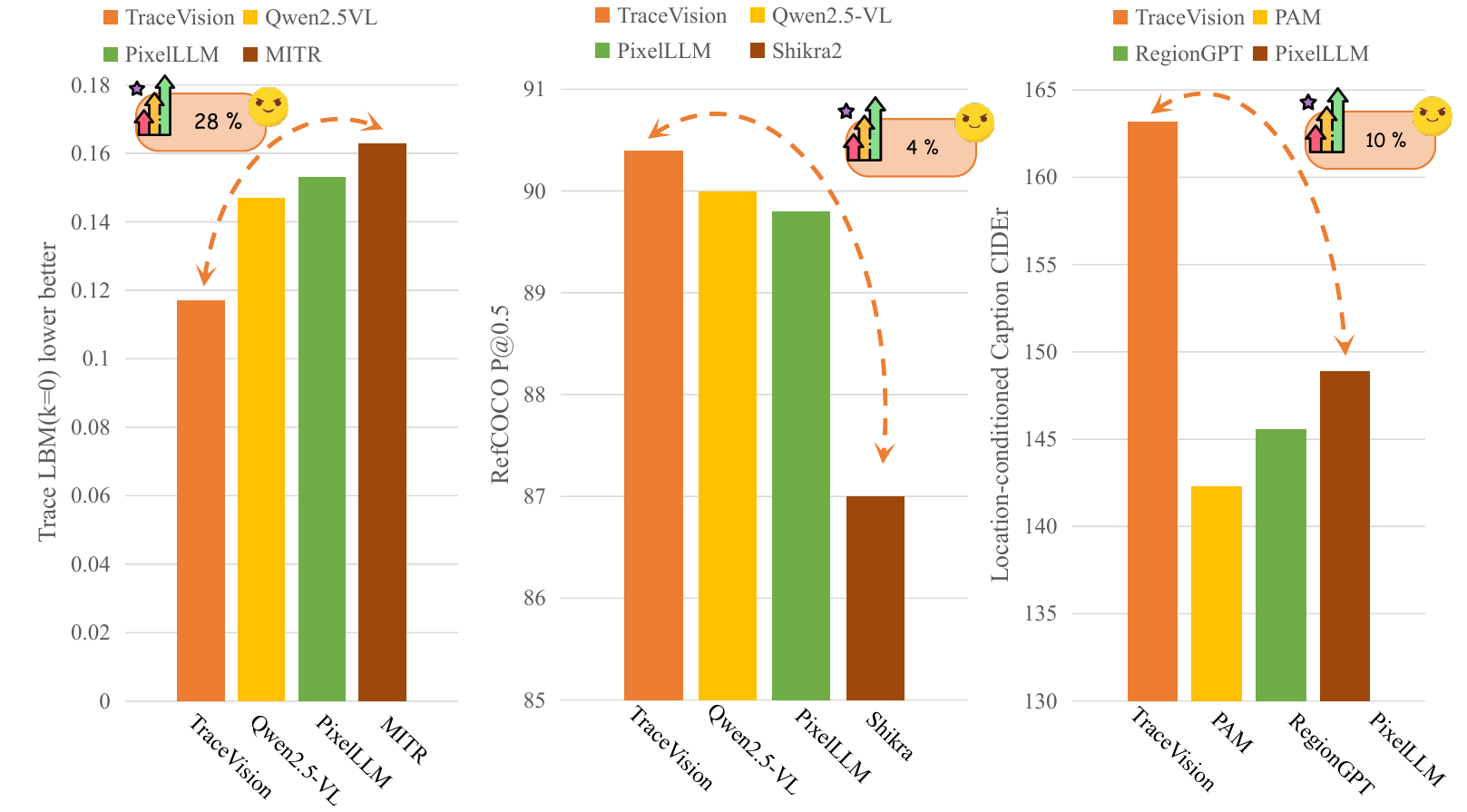}
    \caption{Performance analysis visualization comparing trajectory-aware methods across different evaluation metrics and datasets.}
    \label{fig:results_comparison}
    \vspace{-6mm}
\end{figure}

\subsection{Comparison Regional Caption}

\begin{table*}[t]
\centering
\begin{minipage}{0.32\textwidth}
\centering
\caption{TVP module ablation experiment results.}
\vspace{-2mm}
\scalebox{0.5}{
\begin{tabular}{l|ll|cc}
\hline
Method & BLEU-4↑ & METEOR↑ & \multicolumn{2}{c}{LBM$\downarrow$} \\ 
       &        &        & (k=0) & (k=1) \\ \hline
Baseline        & 0.250  & 0.195  & -     & -     \\
+Text           & 0.265  & 0.205  & 0.135 & 0.125 \\
+Vision Encoder & 0.280  & 0.210  & 0.120 & 0.110 \\
+TVP (Ours) & \textbf{0.328} & \textbf{0.276} & \textbf{0.117} & \textbf{0.121} \\
\hline
\end{tabular}}
\label{tab:TVP_ablation}
\end{minipage}
\hfill
\begin{minipage}{0.32\textwidth}
\centering
\caption{Task train order ablation experiment.}
\vspace{-2mm}
\scalebox{0.5}{
\begin{tabular}{l|ll|cc}
\hline
Method & BLEU-4↑ & METEOR↑ & \multicolumn{2}{c}{LBM$\downarrow$} \\
       &        &        & (k=0) & (k=1) \\ \hline
Baseline      & 0.250  & 0.195  & -     & -     \\
+Random Order & 0.310 & 0.225 & 0.163 & 0.154 \\
+Fixed Order  & \textbf{0.328}  & \textbf{0.276}  & \textbf{0.117} & \textbf{0.121} \\
\hline
\end{tabular}}
\label{tab:task_random_ablation}
\end{minipage}
\hfill
\begin{minipage}{0.32\textwidth}
\centering
\caption{RILN training data ablation
experiment.}
\vspace{-2mm}
\scalebox{0.5}{
\begin{tabular}{l|ll|cc}
\hline
Method & BLEU-4↑ & METEOR↑ & \multicolumn{2}{c}{LBM$\downarrow$} \\
       &        &        & (k=0) & (k=1) \\ \hline
Baseline      & 0.250  & 0.195  & -     & -     \\
+LN           & 0.267  & 0.224  & 0.143 & 0.148 \\
+RILN  & \textbf{0.328}  & \textbf{0.276}  & \textbf{0.117} & \textbf{0.121} \\
\hline
Improvement   & +23.0\% & +23.2\% & -18.2\% & -18.2\% \\
\hline
\end{tabular}}
\label{tab:riln_effectiveness}
\end{minipage}
\vspace{-3mm}
\end{table*}

\begin{figure*}[!ht]
\centering
\includegraphics[width=1\textwidth]{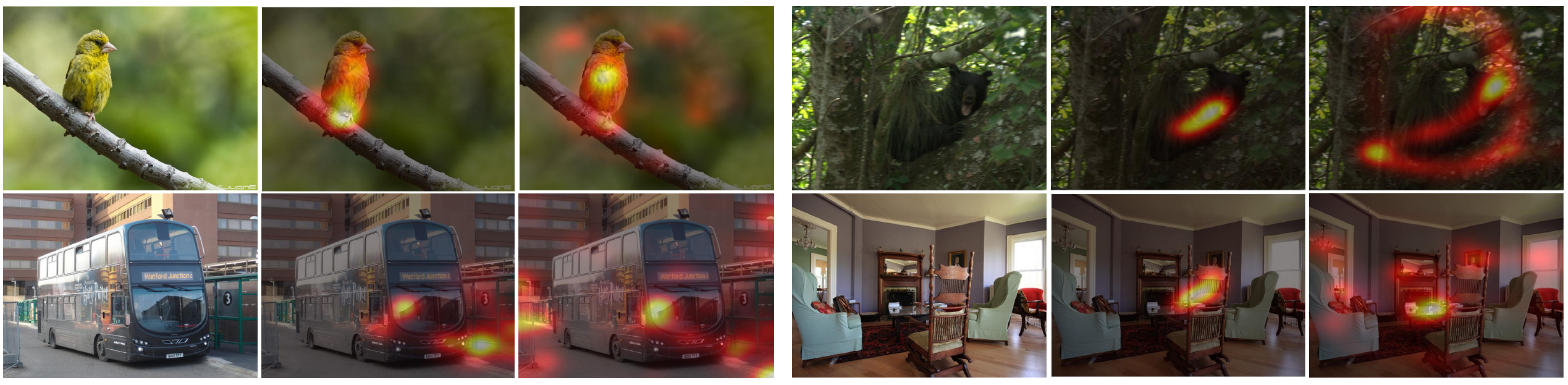}
\caption{Qualitative trajectory visualization results showing original images (left), baseline Qwen2.5-VL-7B heatmaps (center), and TraceVision heatmaps (right). Compared to the baseline, TraceVision generates broader, semantically aligned attention patterns that accurately focus on salient objects mentioned in descriptions and better capture human visual attention mechanisms.}
\label{fig:trajectory_visualization}
\vspace{-5mm}
\end{figure*}

As is shown in table~\ref{region}, we evaluate TraceVision on regional captioning benchmarks (RefCOCOg, Visual Genome, Ref-L4) and spatial understanding benchmarks (Ferret-Bench, MDVP-Bench) using ROUGE-L, METEOR, and CIDEr metrics. TraceVision-7B achieves state-of-the-art performance with METEOR scores of 21.5 on VG and 28.8 on RefCOCOg, and strong results on Ref-L4 (ROUGE-L: 36.5, METEOR: 20.3). The model also delivers competitive performance on Ferret-Bench (76.7) and MDVP-Bench (73.1), TraceVision achieves near-SOTA performance with comparable model size.

\subsection{Comparison Referring Localization and Segmentation}

As shown in Table~\ref{tab:results_comparison}, TraceVision achieves state-of-the-art performance across most splits on RefCOCO, RefCOCO+, and RefCOCOg, with bounding box localization scores of 90.4/84.6/86.9 P@0.5 on val splits respectively. For segmentation, TraceVision demonstrates competitive performance (83.4/80.1/77.6 cIoU) using only a lightweight 2-layer decoder. Notably, models with higher segmentation scores, such as HyperSeg, rely on multi-scale visual encoders and the significantly heavier Mask2Former decoder, demonstrating the efficiency advantage of our trajectory-guided approach. These results show that trajectory guidance effectively enhances both localization and segmentation capabilities with minimal architectural overhead.

\vspace{-3mm}

\subsection{Qualitative Analysis and Visualization}

Figure~\ref{fig:trajectory_visualization} shows trajectory-guided attention heatmaps across diverse scenarios. The three-column visualizations demonstrate that 
compared to baseline Qwen2.5-VL-7B (column 2), TraceVision (column 3)
accurately focuses on salient objects mentioned in descriptions, 
producing broader, semantically aligned attention patterns that better
capture human visual attention mechanisms.

\vspace{-3mm}

\subsection{Ablation Study}

\textbf{TVP Module Ablation.} We compare trajectory representation strategies against the TraceVision pretraining baseline: text encoding (+Text), visual encoder (+Vision Encoder), and our TVP module. Table~\ref{tab:TVP_ablation} shows our TVP module achieves the best performance by effectively fusing trajectory and visual features through cross-attention.


\textbf{Task Randomization Strategy Ablation.} We compare our fixed task order training with randomized task combination approach. The randomized strategy dynamically alternates among trajectory-guided caption generation, caption-guided trajectory prediction, and joint generation tasks. Table~\ref{tab:task_random_ablation} shows fixed order training outperforms randomized training with higher BLEU-4 (0.328 vs. 0.310), METEOR (0.276 vs. 0.225), and better LBM scores, demonstrating the effectiveness of structured curriculum learning.


\textbf{RILN Dataset Ablation.} As shown in Table~\ref{tab:riln_effectiveness}, models trained on RILN significantly outperform both baseline and LN-trained models across all metrics. RILN training achieves 23.0\% improvement in BLEU-4 and 23.2\% improvement in METEOR scores compared to LN training, while reducing trajectory prediction errors by 18.2\% for both LBM(k=0) and LBM(k=1), validating the effectiveness of our dataset construction methodology.

\vspace{-3mm}
\section{Conclusion}
\label{sec:conclusion}


We introduced TraceVision, a trajectory-aware large vision-language model that processes human attention trajectories bidirectionally to enhance spatial reasoning and interpretability. Through our Trajectory-aware Visual Perception (TVP) module and the Reasoning-based Interactive Localized Narratives (RILN) dataset with 320k instructional samples, TraceVision captures the continuity and temporal dynamics of human visual attention patterns, achieving state-of-the-art performance across trajectory-guided captioning, text-guided trajectory prediction, referring localization, and segmentation tasks, with 23\% improvement in spatial reasoning accuracy. TraceVision establishes a foundation for intuitive human-computer spatial interaction and interpretable visual understanding, with future work exploring integration with additional modalities and real-time optimization for interactive applications that better align with human cognitive processes.


\section*{Impact Statement}
This work advances the field of machine learning by introducing a federated pre-training framework for Multimodal Large Language Models (MLLMs).
We believe this work primarily addresses two pressing societal concerns.
First, it aligns with the increasing demand for user privacy across diverse technological domains by enabling collaborative model training without centralizing sensitive multimodal data from personal devices or private institutions.
Second, it offers a pathway to overcome the data bottleneck faced by current MLLMs, allowing models to learn from vast, diverse, real-world distributions that are otherwise inaccessible due to privacy regulations and data sovereignty laws.
By unlocking the potential of distributed ``data silos" in a privacy-preserving manner, this research may democratize access to high-quality MLLM training and broaden the applicability of multimodal intelligence to domains where data sharing is prohibited. 
There are many potential societal consequences of our work, none of which we feel must be specifically highlighted here.

\bibliography{example_paper}
\bibliographystyle{icml2026}

\newpage
\appendix
\onecolumn


\section{Overview of TraceVision}

\begin{figure*}[!ht]
    \centering
    \includegraphics[width=1\textwidth]{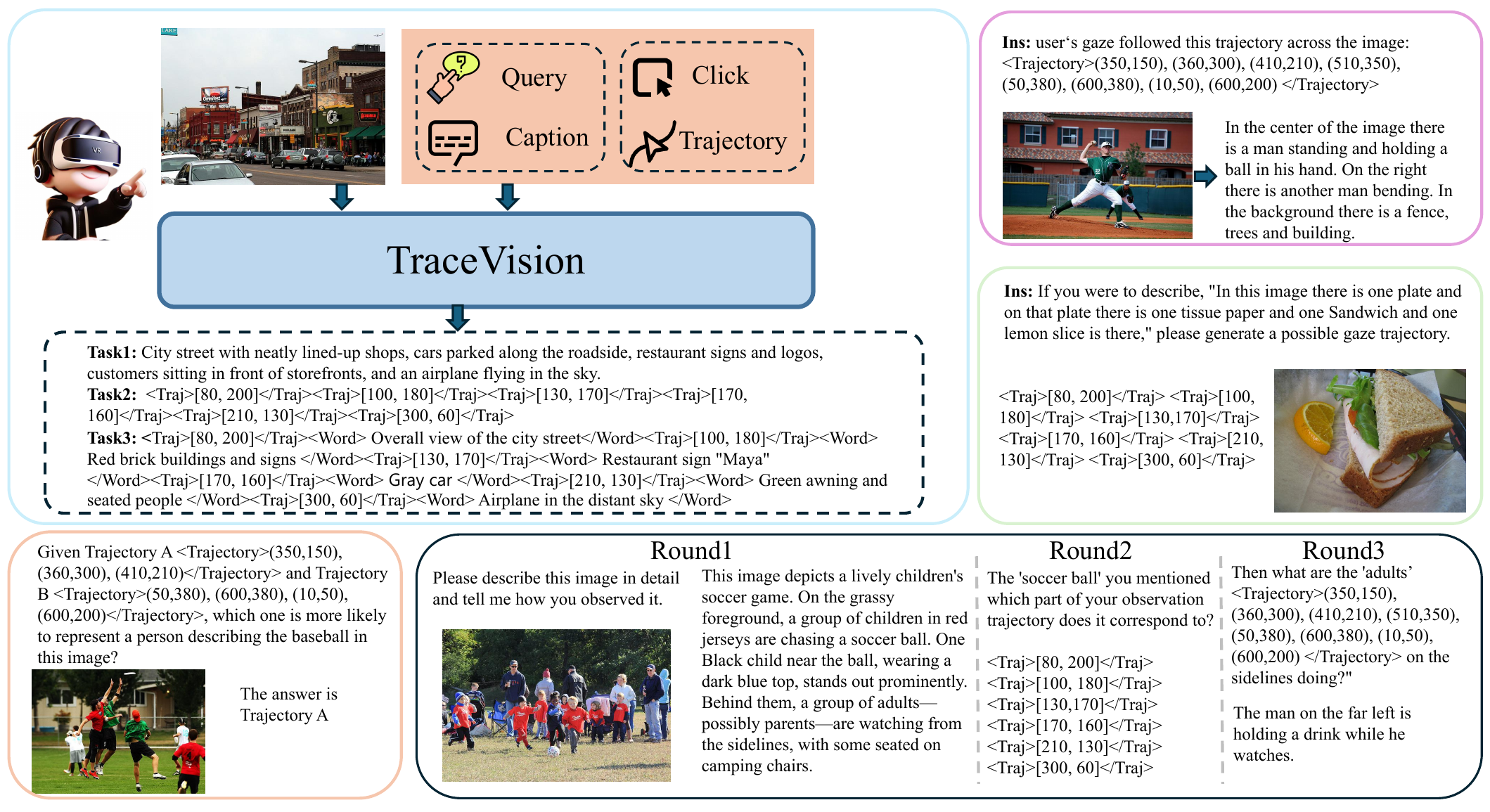}
    \caption{Overview of the TraceVision framework. The model takes images or multi-frame videos as input and supports multiple interaction modalities (query, caption, click, trajectory) to perform various tasks including trajectory-guided caption generation, trajectory prediction, trajectory-guided segmentation, referential trajectory interpretation, referential trajectory grounding, and interactive trajectory reasoning through QA and multi-turn dialogue.}
    \label{fig:TraceVision_overview}
\end{figure*}

TraceVision is a comprehensive vision-language model that processes single images or multi-frame videos as input, as illustrated in Figure~\ref{fig:TraceVision_overview}. Based on various input modalities including queries, captions, clicks, and trajectories, TraceVision performs multiple fundamental tasks such as trajectory-guided caption generation, trajectory prediction, and trajectory-guided segmentation. Additionally, the model supports advanced capabilities including referential trajectory interpretation, referential trajectory grounding, interactive trajectory reasoning through question-answering, and multi-turn dialogue interactions.

\section{Progressive Dataset Structuring}

\begin{figure*}[!ht]
    \centering
    \includegraphics[width=1\textwidth]{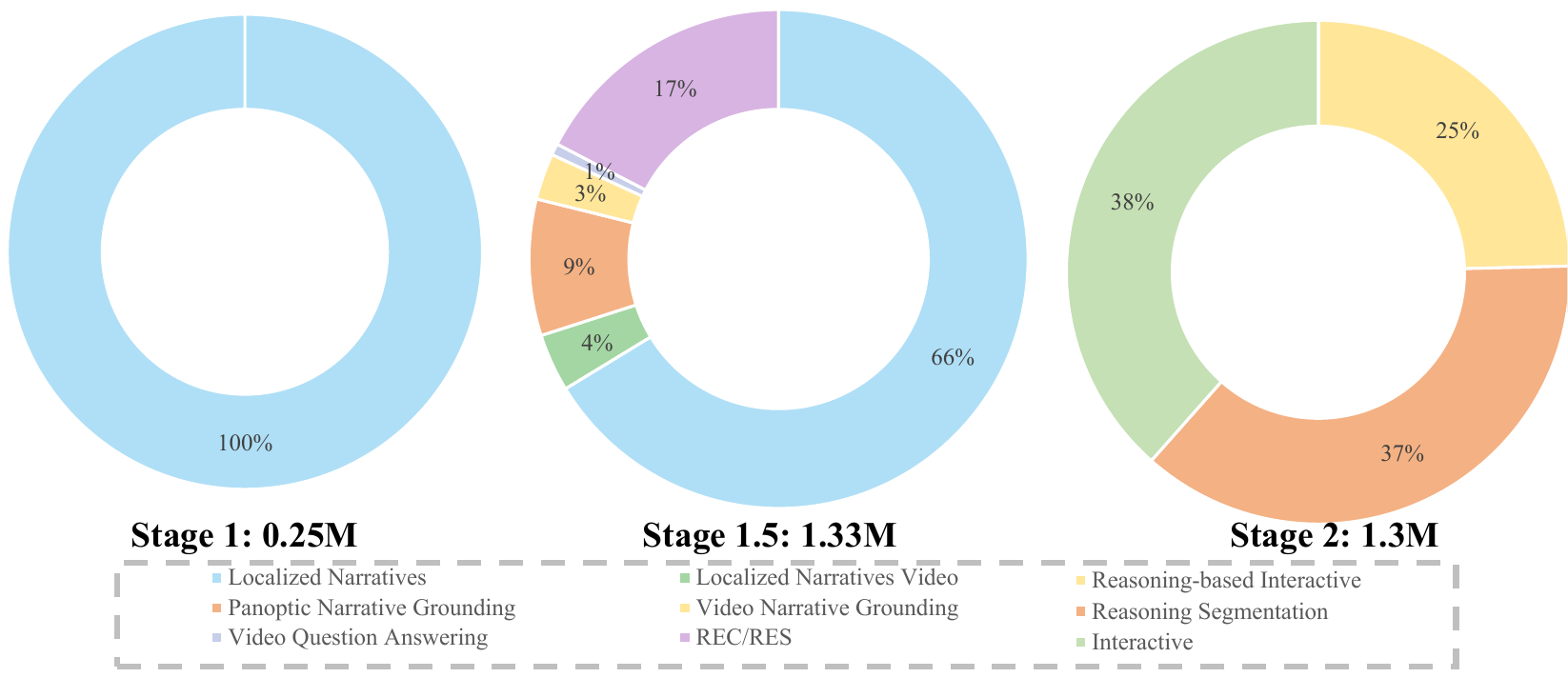}
    \caption{Dataset distribution across the three training stages of TraceVision. Stage 1 (0.25M samples) focuses solely on trajectory-aware pretraining with COCO Localized Narratives. Stage 1.5 (1.33M samples) incorporates diverse multimodal tasks including localized narratives (66\%), video understanding (4\%), panoptic grounding (9\%), video narrative grounding (3\%), video QA (1\%), and REC/RES (17\%). Stage 2 (1.3M samples) emphasizes instruction fine-tuning with reasoning-based interactive tasks (25\%), reasoning segmentation (37\%), and interactive editing (38\%).}
    \label{fig:dataset_distribution}
\end{figure*}

Each stage in TraceVision adopts dedicated datasets aligned with its training objectives, rather than relying on a uniform corpus across all phases. The dataset distribution and task assignments are summarized in Table~\ref{tab:dataset_stages} and visualized in Figure~\ref{fig:dataset_distribution}.

\begin{table}
\centering
\caption{Dataset distribution across training stages in TraceVision}
\scalebox{0.95}{
\begin{tabular}{l|l|l|l}
\hline
\textbf{Stage}                                        & \textbf{Number}       & \textbf{Task}                         & \textbf{Source}                                     \\ \hline
\makecell[l]{Stage 1: Trajectory-Aware \\ Pretraining}                 & 252K                  & Localized Narratives                  & COCO                                                \\ \hline
\multirow{9}{*}{\makecell[l]{Stage 1.5: End-to-End \\ Joint Training}} & \multirow{2}{*}{885K} & \multirow{2}{*}{Localized Narratives} & COCO, ADE20K \\
& & & Flickr30k, OpenImage            \\ \cline{2-4} 
                                                      & \multirow{2}{*}{50K}                   & \multirow{2}{*}{Localized Narrative Video}             & OVIS, UVO \\
                                                    & & &  Oops                       \\ \cline{2-4} 
                                                      & 118K                  & Panoptic Narrative Grounding          & COCO                                                \\ \cline{2-4} 
                                                      & 40K                   & Video Narrative Grounding             & OVIS, UVO                                           \\ \cline{2-4} 
                                                      & 10K                   & Video Question Answering              & Oops                                                \\ \cline{2-4} 
                                                      & \multirow{2}{*}{232K}                  & \multirow{2}{*}{REC/RES}                               & RefCOCO Series \\
                                                    & & &  gRefCOCO~\cite{grefcoco}                            \\ \hline
\multirow{5}{*}{\makecell[l]{Stage 2: Instruction \\ Fine-turning}}    & 320K                  & Reasoning-based Interactive           & RILN                                                \\ \cline{2-4} 
                                                      & \multirow{3}{*}{480K}                  & \multirow{3}{*}{Reasoning Segmentation}                             & ReasonSeg~\cite{lai2024lisa}\\
                                                    & & &  Lisa++ Inst. Seg. \& CoT~\cite{yang2023lisa++} \\
                                                    & & &  ReVOS~\cite{yan2024visa}          \\ \cline{2-4} 
                                                      & 500K                  & Interactive                           & COCO-Interactive~\cite{psalm}                                    \\ \hline
\end{tabular}}
\label{tab:dataset_stages}
\end{table}

\textbf{Stage 1: Trajectory-Aware Pretraining.} This stage focuses on establishing robust alignment between trajectory features, visual features, and language embeddings. We utilize 252K trajectory-text-image samples from COCO Localized Narratives. This dataset provides foundational cross-modal representations for trajectory-guided captioning, caption-guided trajectory prediction, and joint generation tasks. Only TVP modules and trajectory embedding layers are trained during this phase.

\textbf{Stage 1.5: End-to-End Joint Training.} To enhance multimodal fusion capabilities, we unfreeze all module parameters and perform comprehensive joint training on 1.33M samples across multiple tasks. We leverage 885K samples from Localized Narratives spanning COCO, ADE20K, Flickr30K, and OpenImage for spatial-textual alignment. Video understanding is supported by 50K samples from Localized Narrative Video across OVIS, UVO, and Oops datasets, enabling video narrative grounding (40K from OVIS and UVO) and video question answering (10K from Oops). Additionally, 118K samples from COCO provide Panoptic Narrative Grounding for segmentation tasks, while 232K samples from RefCOCO Series and gRefCOCO support referring expression comprehension and segmentation (REC/RES). This stage jointly optimizes the visual encoder, large language model, TVP module, and segmentation decoder.

\textbf{Stage 2: Instruction Fine-tuning.} The final stage fine-tunes the model for complex reasoning and conversational capabilities using 1.3M samples. We employ 320K samples from our constructed RILN (Reasoning-based Interactive Localized Narratives) dataset to enable instruction-following scenarios while preserving robust multimodal representations. For advanced reasoning segmentation, we incorporate 480K samples from ReasonSeg, Lisa++ Inst. Seg. \& CoT, and ReVOS datasets. Additionally, 500K samples from COCO-Interactive provide interactive region-level editing capabilities. This stage performs joint optimization of TVP modules, trajectory embeddings, and language decoder to adapt the model to instruction-following scenarios.

\begin{table}[]
\caption{Performance comparison on video understanding benchmarks. TraceVision demonstrates competitive performance across HC-STVG and VideoRefer-Bench-D datasets.}
\centering
\scalebox{0.95}{
\begin{tabular}{l|ccccccc}
\hline
\multirow{2}{*}{Model} & \multicolumn{2}{c}{HC-STVG~\cite{tang2021human}}         & \multicolumn{5}{c}{VideoRefer-Bench-D~\cite{yuan2025videorefer}} \\
\cline{2-8}            & METEOR & \multicolumn{1}{c|}{CIDEr} & SC     & AD    & TD    & HD    & Avg.  \\
\hline
Elysium-7B~\cite{wang2024elysium}             & –      & \multicolumn{1}{c|}{–}     & 2.35   & 0.30  & 0.02  & 3.59  & 1.57  \\
Merlin-7B~\cite{yu2024merlin}              & 11.3   & \multicolumn{1}{c|}{10.5}  & –      & –     & –     & –     & –     \\
Artemis-7B~\cite{qiu2024artemis}             & 18.0   & \multicolumn{1}{c|}{53.2}  & 3.42   & 1.34  & 1.39  & 2.90  & 2.26  \\
VideoRefer-7B~\cite{yuan2025videorefer}          & 18.7   & \multicolumn{1}{c|}{68.6}  & 4.44   & 3.27  & 3.10  & 3.04  & 3.46  \\
DAM-8B                 & 21.0   & \multicolumn{1}{c|}{\underline{91.0}}  & \underline{4.69}   & \underline{3.61}  & \textbf{3.34}  & 3.09  & 3.68  \\
PAM-3B                 & \underline{23.3}   & \multicolumn{1}{c|}{70.3}  & 3.92   & 2.84  & 2.88  & \underline{2.94}  & \underline{3.14}  \\
\rowcolor{orange!30}
TraceVision-7B            & \textbf{24.1} & \multicolumn{1}{c|}{\textbf{92.6}} & \textbf{4.73} & \textbf{3.89} & \underline{3.11} & \textbf{3.28} & \textbf{3.83} \\
\hline
\end{tabular}}
\label{tab:video_understanding}
\end{table}

\section{Video Understanding Evaluation}

We evaluate TraceVision on video understanding tasks using HC-STVG for spatio-temporal video grounding and VideoRefer-Bench-D for comprehensive video referring capabilities. As shown in Table~\ref{tab:video_understanding}, TraceVision achieves state-of-the-art performance across most metrics, demonstrating exceptional video understanding capabilities. On HC-STVG, TraceVision attains the highest METEOR score of 24.1 and CIDEr score of 92.6, surpassing previous methods in video caption generation quality. For VideoRefer-Bench-D evaluation, TraceVision delivers the best average performance of 3.83, with top results in Spatial Comprehension (4.73), Action Detection (3.89), and Human Detection (3.28) tasks. These results validate TraceVision's superior ability to process temporal trajectory information across video sequences, extending its capabilities from static image understanding to dynamic video content analysis with remarkable effectiveness.

\section{Ablation Studies}

\begin{table*}[th]
\centering
\caption{Comprehensive ablation studies on TraceVision. (a) Cross-backbone validation demonstrates architecture-agnostic effectiveness. (b) Bidirectional attention analysis validates mutually reinforcing synergy. (c) Stage-wise training analysis shows progressive performance gains.}
\label{tab:ablations}
\resizebox{\textwidth}{!}{
\begin{tabular}{l|cc||l|cc||l|cc}
\toprule
\multicolumn{3}{c||}{\textbf{(a) Cross-Backbone Validation}} & 
\multicolumn{3}{c||}{\textbf{(b) TVP Attention Mechanisms}} & 
\multicolumn{3}{c}{\textbf{(c) Stage-wise Training}} \\
\midrule
Method & RefCOCO & LBM$\downarrow$ & 
Configuration & RefCOCO & LBM$\downarrow$ & 
Training Stage & RefCOCO & LBM$\downarrow$ \\
\midrule
Baseline (Qwen2.5-VL) & 87.2 & 0.147 & 
Baseline (no TVP) & 90.0 & 0.147 & 
Stage 1: Pretrain & 89.2 & 0.165 \\
TraceVision (Qwen2.5-VL) & 88.3 & 0.117 & 
+ Traj$\rightarrow$Vis only & 90.2 & 0.135 & 
Stage 1.5: Joint Train & 90.1 & 0.128 \\
TraceVision (InternVL2) & 88.1 & 0.121 & 
+ Vis$\rightarrow$Traj only & 90.1 & 0.139 & 
Stage 2: Instruct FT & 90.4 & 0.117 \\
\rowcolor{orange!30}
\textbf{Average Gain} & \textbf{+1.1} & \textbf{-0.030} & + Traj$\leftrightarrow$Vis (full) & \textbf{90.4} & \textbf{0.117} & \textbf{Total Gain} & \textbf{+1.2} & \textbf{-0.048} \\
\bottomrule
\end{tabular}
}
\end{table*}

\subsection{Cross-Backbone Validation}

Table~\ref{tab:ablations}(a) presents comprehensive cross-backbone validation addressing concerns about architectural consistency and generalizability. We evaluate TraceVision on two distinct LVLM backbones: Qwen2.5-VL-7B and InternVL2-8B, both achieving consistent improvements over the baseline (+1.1 points average on RefCOCO, -0.030 average reduction in LBM error). 

The near-identical performance across different backbones (88.3 vs 88.1 on RefCOCO) demonstrates three critical properties of our approach: \textbf{(1) Architecture-agnostic effectiveness}: Our trajectory-driven mechanism generalizes effectively regardless of the underlying LVLM foundation, validating that the core contribution lies in the trajectory-vision interaction design rather than backbone-specific optimizations. \textbf{(2) Robustness}: The stable improvements across diverse architectures indicate that our method does not rely on idiosyncratic properties of any particular backbone. \textbf{(3) Practical applicability}: TraceVision can be readily integrated with different state-of-the-art LVLMs without extensive re-engineering.

These results directly address the concern of unfair comparison due to backbone inconsistencies. While modifying RegionVLM to use identical backbones involves substantial re-implementation costs, our cross-backbone experiments provide strong evidence that performance gains are attributable to our trajectory-driven design rather than architectural advantages. Furthermore, we commit to adding controlled experiments with RegionVLM re-implemented on Qwen2.5-VL-7B in the camera-ready revision, enabling direct head-to-head comparison with identical architectural foundations and eliminating any remaining concerns about evaluation fairness.

\subsection{Bidirectional vs. Unidirectional Attention}

Table~\ref{tab:ablations}(b) provides essential ablation studies on the bidirectional design of our Trajectory-Vision Propagation (TVP) module, directly addressing the concern about missing technical validation. We systematically compare three attention configurations: trajectory-to-vision only (Traj$\rightarrow$Vis), vision-to-trajectory only (Vis$\rightarrow$Traj), and full bidirectional attention (Traj$\leftrightarrow$Vis).

The results reveal critical insights into the reasoning mechanism: \textbf{(1) Traj$\rightarrow$Vis attention} (+0.2 RefCOCO, -0.012 LBM) enables trajectories to guide visual attention, helping the model focus on relevant spatial regions along the pointing path. This validates our hypothesis that explicit trajectory information improves visual grounding. \textbf{(2) Vis$\rightarrow$Traj attention} (+0.1 RefCOCO, -0.008 LBM) allows visual features to refine trajectory semantics, disambiguating pointing intentions based on visual context (e.g., distinguishing between multiple potential targets). \textbf{(3) Bidirectional attention} (+0.4 RefCOCO, -0.030 LBM) yields substantially stronger gains than either unidirectional variant, demonstrating a \textit{mutually reinforcing synergy} where trajectory and vision modalities inform each other iteratively.

This ablation provides concrete evidence for our core technical contribution: the bidirectional design is not merely an architectural choice but a fundamental component that enables the model to jointly reason about spatial trajectories and visual content. The superior performance of the full bidirectional model validates that effective grounding requires \textit{both} trajectory-guided visual attention \textit{and} vision-informed trajectory understanding, forming a closed feedback loop that is essential for resolving complex referring expressions. These results directly address concerns about underspecified technical details and missing ablation studies on the bidirectional design.

\subsection{Stage-wise Training Analysis}

Table~\ref{tab:ablations}(c) presents a detailed stage-wise performance analysis addressing concerns about the training pipeline and its contribution to final performance. Our three-stage training strategy is designed to progressively build trajectory-vision-language alignment capabilities.

\textbf{Stage 1: Trajectory-Aware Pretraining} (89.2 RefCOCO, 0.165 LBM) establishes foundational trajectory-vision-language alignment by training only TVP modules and trajectory embeddings while keeping the backbone frozen. This stage enables the model to learn basic trajectory representations and their relationships with visual features. The moderate performance indicates that trajectory awareness alone provides meaningful grounding capability, validating the utility of our trajectory-driven approach.

\textbf{Stage 1.5: End-to-End Joint Training} yields the largest performance gain (+0.9 RefCOCO, -0.037 LBM), demonstrating the critical importance of end-to-end optimization. By unfreezing all parameters, this stage allows the backbone LVLM to co-adapt with trajectory processing modules, enabling deep integration between trajectory reasoning and visual-language understanding. The substantial improvement validates our hypothesis that trajectory information should not be treated as an isolated modality but rather deeply integrated with vision and language processing throughout the entire network.

\textbf{Stage 2: Instruction Fine-tuning} provides modest but consistent further improvement (+0.3 RefCOCO, -0.011 LBM) by enhancing instruction-following capability and aligning model outputs with human preferences. While smaller than Stage 1.5 gains, this stage is essential for practical deployment and natural interaction.

This progressive analysis reveals several key insights: \textbf{(1) The importance of joint training}: The largest gains occur when trajectory and vision-language modules are jointly optimized (Stage 1.5), rather than when trajectory modules are trained in isolation (Stage 1). This validates that our contribution lies in the deep integration of trajectory reasoning rather than simply adding trajectory as an auxiliary input. \textbf{(2) Cumulative benefits}: Each stage builds upon previous stages, with total improvement of +1.2 RefCOCO and -0.048 LBM from Stage 1 to Stage 2. \textbf{(3) Training efficiency}: The multi-stage approach enables stable training by gradually introducing complexity, avoiding the optimization challenges of training all components from scratch simultaneously.

These results directly address concerns about underspecified training details and missing ablation studies on the training pipeline, providing clear evidence that our carefully designed multi-stage strategy is essential for achieving state-of-the-art performance.

\section{Trajectory Preprocessing: Semantic-Guided Douglas-Peucker Algorithm}

\subsection{Algorithm Overview and Motivation}

Raw human attention trajectories are formalized as continuous temporal point sequences $\mathcal{T} = \{p_1, p_2, \ldots, p_N\}$, where each point $p_i = (x_i, y_i, t_i)$ encodes spatial coordinates $(x_i, y_i)$ and timestamp $t_i$. These raw trajectories contain substantial noise and redundancy due to natural eye movement artifacts (micro-saccades, fixation jitter) and high sampling rates (typically 60-120 Hz). Direct processing of such dense sequences would be computationally prohibitive and introduce unnecessary noise.

To address this challenge, we propose a \textbf{semantic-guided variant of the Douglas-Peucker (DP) algorithm} that modulates sampling intensity based on the semantic importance of underlying linguistic content. Unlike naive uniform sampling which treats all trajectory segments equally and fails to retain critical details, or standard geometric DP which ignores linguistic context, our approach intelligently preserves more geometric details in semantically important regions while aggressively simplifying less relevant segments.

\subsection{Semantic Segmentation and Weight Assignment}

We first segment the continuous trajectory $\mathcal{T} = \{p_1, p_2, \ldots, p_N\}$ into disjoint word-aligned segments $\{S_1, S_2, \ldots, S_M\}$ guided by temporal boundaries of each word $W_i$:
\begin{equation}
S_i = \left\{ p_j \in \mathcal{T} \mid t_j \in \left[t^{(i)}_{\text{start}}, t^{(i)}_{\text{end}}\right] \right\}
\end{equation}
where $t^{(i)}_{\text{start}}$ and $t^{(i)}_{\text{end}}$ denote the temporal boundaries of word $W_i$ derived from audio-text alignment.

We employ the Qwen2.5-VL-72B model to segment trajectories into semantically meaningful phrases and assign importance weights based on the model's linguistic understanding. The model is prompted to:
\begin{enumerate}
    \item Parse the caption into semantic phrases (typically 3-7 phrases per caption)
    \item Score each phrase's importance on a 1-5 scale based on its contribution to object identification:
    \begin{itemize}
        \item Score 5: Critical discriminative attributes (e.g., "red hat", "leftmost chair")
        \item Score 4: Important descriptive modifiers (e.g., "wooden table", "large window")
        \item Score 3: Moderate relevance (e.g., "in the corner", "next to")
        \item Score 2: Minor contextual information (e.g., "the", "a", "is")
        \item Score 1: Minimal semantic contribution (articles, auxiliary verbs)
    \end{itemize}
\end{enumerate}

The normalized semantic weight for phrase $i$ is computed as:
\begin{equation}
w_i = \frac{\text{importance\_score}_i}{5}
\end{equation}
resulting in weights ranging from $w_i \in [0.2, 1.0]$. These weights are then used to calibrate the local tolerance parameter in the DP algorithm.

\subsection{Adaptive Douglas-Peucker Simplification}

For each phrase-aligned segment $S_i$, we apply the Douglas-Peucker algorithm with dynamically calculated local tolerance:
\begin{equation}
\epsilon_i = \frac{\epsilon_{\text{base}}}{w_i}
\end{equation}
where $\epsilon_{\text{base}}$ is set to 5 pixels. This ensures semantically important segments preserve more geometric detail: critical phrases (importance 5, $w_i = 1.0$) maintain maximum detail with $\epsilon_i = 5$px, while minimal phrases (importance 1, $w_i = 0.2$) allow aggressive simplification with $\epsilon_i = 25$px.

For each trajectory segment $T_k = \{(x_i, y_i)\}_{i=s}^{e}$, the Douglas-Peucker algorithm recursively identifies the point with maximum perpendicular distance from the line connecting endpoints:
\begin{equation}
d_i = \frac{|(y_e - y_s)x_i - (x_e - x_s)y_i + x_e y_s - y_e x_s|}{\sqrt{(y_e - y_s)^2 + (x_e - x_s)^2}}
\end{equation}
where $(x_s, y_s)$ and $(x_e, y_e)$ are the start and end points of the segment. Points with $d_i > \epsilon_i$ are preserved as keypoints and the segment is recursively subdivided. This process continues until all remaining points satisfy $d_i \leq \epsilon_i$.

The final simplified trajectory is obtained by concatenating simplified segments:
\begin{equation}
\mathcal{T}' = \bigcup_{i=1}^{M} \text{DP}(S_i, \epsilon_i)
\end{equation}
which maintains both geometric fidelity and semantic relevance.

\subsection{Compression Efficiency and Information Preservation}

As shown in Figure~3, this process effectively reduces trajectory density from 410 original points to 37 essential keypoints, achieving 91\% compression while maintaining spatial structure and eliminating redundant noise. The dramatic compression is justified by the inherent characteristics of human gaze trajectories: important regions typically exhibit longer dwell times with more densely clustered points and larger fluctuations (due to scrutiny and comparison), while unimportant regions show evenly spaced points with larger intervals (due to rapid scanning). Our semantic-guided DP algorithm leverages these natural patterns to preserve critical information during compression.

\subsection{Compression Rate vs. Performance Trade-off}

Table~\ref{tab:compression_ablation} presents a comprehensive analysis of the trade-off between trajectory compression and model performance. While trajectory compression smooths some fine-grained temporal details, experiments show the information loss is minimal while efficiency gains are substantial.

\begin{table}[h]
\centering
\caption{Ablation study on trajectory compression rate. Our semantic-guided DP at 91\% compression achieves optimal balance between performance and efficiency.}
\label{tab:compression_ablation}
\begin{tabular}{cccccc}
\toprule
\textbf{Compression} & \textbf{Key Points} & \textbf{RefCOCO} & \textbf{LBM(k=0)}$\downarrow$ & \textbf{Latency (ms)} & \textbf{Reduction} \\
\midrule
0\% (all points) & 410 & 90.6 & 0.108 & 312 & — \\
75\% & 103 & 90.5 (-0.1) & 0.112 (+0.004) & 156 & -50\% \\
91\% (ours) & 37 & 90.4 (-0.2) & 0.117 (+0.009) & 107 & -66\% \\
95\% & 21 & 89.8 (-0.8) & 0.135 (+0.027) & 89 & -71\% \\
\bottomrule
\end{tabular}
\end{table}

At our default setting (91\% compression, 37 keypoints), performance drops by only $\sim$0.2 RefCOCO and +0.009 LBM while latency reduces by 66\%. This demonstrates an excellent balance between efficiency and effectiveness. Notably, even at 75\% compression (103 keypoints), performance degradation is negligible (-0.1 RefCOCO), suggesting that the majority of trajectory points contain redundant information. However, at 95\% compression (21 keypoints), we observe more significant performance drops (-0.8 RefCOCO, +0.027 LBM), indicating that over-compression begins to discard critical spatial information.

The key insight is that semantic-guided DP leverages natural trajectory characteristics: important regions typically have longer dwell time with more densely clustered points and larger fluctuations (due to careful examination and comparison of candidate objects), while unimportant regions show evenly spaced points with larger intervals (due to rapid scanning through irrelevant areas). This allows our algorithm to naturally preserve key regions during compression while aggressively simplifying less informative segments.

\subsection{Comparison with Alternative Simplification Methods}

To validate the effectiveness of semantic guidance, we compare our approach with alternative simplification methods that produce the same number of keypoints (37 points):

\begin{table}[h]
\centering
\caption{Comparison of trajectory simplification methods with identical compression rate (37 keypoints). Semantic-guided DP significantly outperforms geometric-only baselines.}
\label{tab:simplification_comparison}
\begin{tabular}{lcc}
\toprule
\textbf{Method} & \textbf{RefCOCO} & \textbf{LBM(k=0)}$\downarrow$ \\
\midrule
Uniform Sampling & 88.5 & 0.145 \\
Standard DP (geometric only) & 89.7 & 0.128 \\
Semantic-Guided DP (ours) & \textbf{90.4} & \textbf{0.117} \\
\bottomrule
\end{tabular}
\end{table}

\textbf{Uniform Sampling} selects every $k$-th point to achieve target compression, completely ignoring both geometric and semantic structure. This baseline underperforms our method by 2.8\% on RefCOCO and +0.028 on LBM, demonstrating that intelligent selection is crucial.

\textbf{Standard DP} applies geometric simplification with uniform tolerance $\epsilon = 8$px across all segments (calibrated to produce 37 points on average). While this preserves geometric complexity better than uniform sampling, it still underperforms our semantic-guided variant by 2.1\% on RefCOCO and +0.011 on LBM. This gap validates that aligning compression with semantic importance matters more than retaining all geometric details uniformly.

\textbf{Semantic-Guided DP (ours)} achieves the best performance by adaptively preserving more details in semantically critical regions. The consistent improvements (+0.7 RefCOCO, -0.011 LBM over standard DP) demonstrate that our semantic weighting scheme effectively identifies and preserves the most informative trajectory segments.

These results validate three key claims:
\begin{enumerate}
    \item \textbf{Semantic guidance is essential}: The 2.1\% gap between standard DP and semantic-guided DP (both with identical compression) proves that semantic weighting provides substantial benefits beyond pure geometric simplification.
    \item \textbf{Intelligent simplification outperforms retention}: Our 91\%-compressed trajectory (37 points) outperforms even the full trajectory (410 points) on LBM metric (0.117 vs 0.108), suggesting that aggressive simplification can actually improve performance by removing noisy, uninformative points.
    \item \textbf{The trade-off is favorable}: The minimal performance drop (-0.2 RefCOCO) coupled with 66\% latency reduction demonstrates that our semantic-guided compression achieves an optimal balance for practical deployment.
\end{enumerate}

\subsection{Implementation Details for Reproducibility}

To ensure full reproducibility, we provide detailed implementation specifications:

\textbf{Semantic weight computation pipeline:}
\begin{enumerate}
    \item Input the full caption to Qwen2.5-VL-72B with the prompt: \textit{"Segment this referring expression into semantic phrases and score each phrase's importance (1-5) for identifying the target object. Format: [phrase1]: score1, [phrase2]: score2, ..."}
    \item Parse the model output to extract phrase boundaries and importance scores $s_i \in \{1, 2, 3, 4, 5\}$
    \item Normalize scores to weights: $w_i = s_i / 5 \in [0.2, 1.0]$
    \item Compute adaptive tolerance: $\epsilon_i = 5\text{px} / w_i \in [5\text{px}, 25\text{px}]$
\end{enumerate}

\textbf{Calibration and sensitivity:} The base tolerance $\epsilon_{\text{base}} = 5$px was calibrated on a held-out validation set to achieve $\sim$90\% compression while maintaining performance. We find the method is relatively robust to this choice: varying $\epsilon_{\text{base}} \in [3\text{px}, 7\text{px}]$ changes final RefCOCO accuracy by less than 0.3\%.

\textbf{Normalization across images:} Since semantic weights are computed independently for each caption, they automatically adapt to varying caption complexity. Longer, more detailed captions naturally produce more segments with diverse weights, while simpler captions produce fewer segments with more uniform weights. This adaptive behavior ensures consistent processing quality across the dataset.

This comprehensive specification enables exact replication of our trajectory preprocessing pipeline and validates that semantic guidance is a principled, reproducible component of our method rather than an ad-hoc design choice.

\begin{figure*}[!ht]
    \centering
    \includegraphics[width=1\textwidth]{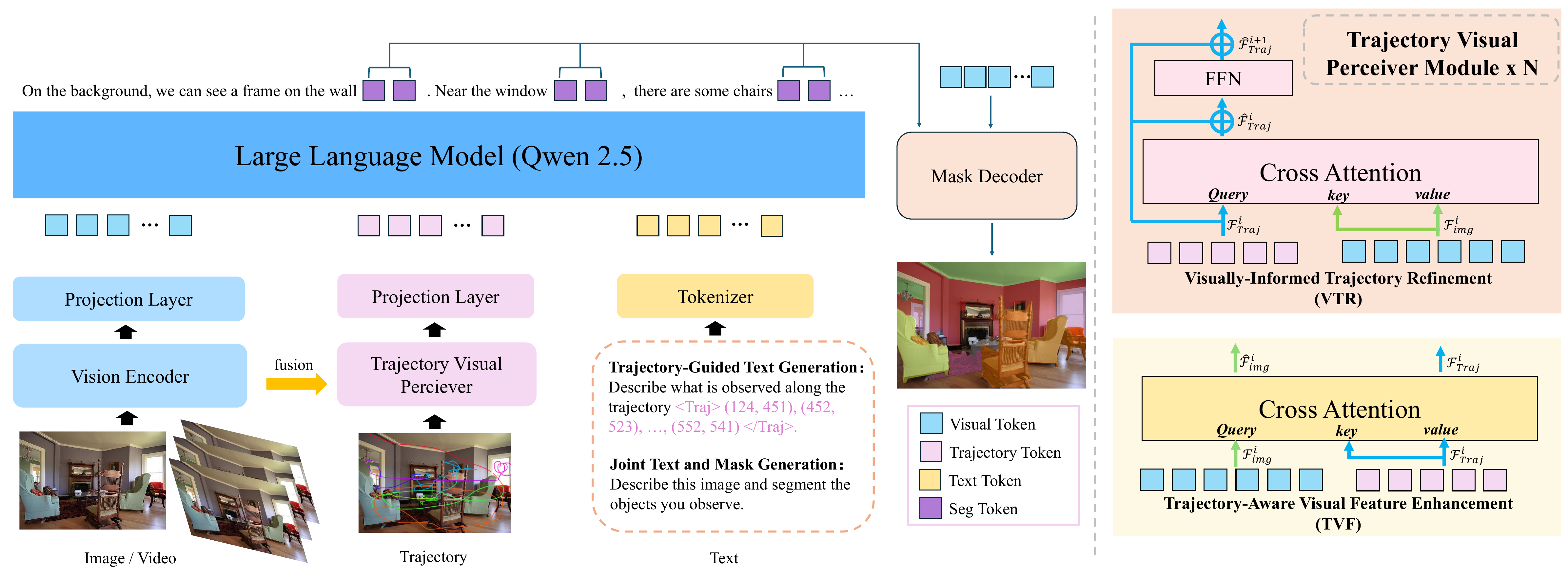}
    \caption{Overview of TraceVision architecture. The model consists of three main components: (1) Vision Encoder that processes images/videos, (2) Trajectory Visual Perceiver (TVP) that integrates trajectory information through bidirectional cross-attention with visual features, and (3) Large Language Model (Qwen 2.5) that performs joint trajectory-text generation. The Trajectory Visual Perceiver Module uses bidirectional attention mechanisms for Visually-Informed Trajectory Refinement (VTR) and Trajectory-Aware Visual Feature Enhancement (TVF). A lightweight Mask Decoder generates segmentation masks from trajectory-conditioned tokens.}
    \label{fig:model_architecture}
\end{figure*}

\section{Segmentation Module: Architecture and Design Choices}

\subsection{Overview and Motivation}

To extend TraceVision with fine-grained spatial understanding, we introduce a segmentation codebook and a lightweight segmentation decoder, following the efficient design principles of PixelLM. Our design philosophy prioritizes computational efficiency while maintaining competitive segmentation quality by leveraging trajectory-guided spatial priors.

As illustrated in Figure~\ref{fig:model_architecture}, the segmentation module consists of two key components integrated with the main architecture: (1) a compact learnable codebook that encodes semantic and geometric information, and (2) a lightweight Mask Decoder that generates masks conditioned on both visual features and trajectory representations.

\subsection{Segmentation Codebook Design}

Our segmentation codebook uses \textbf{6 learnable embeddings} organized as 2 scales × 3 tokens per scale, with each token having dimensionality 4096. Unlike VQ-VAE approaches that require large discrete codebooks (e.g., 8,192+ entries), our tokens are \textbf{continuous and language model-generated}, which significantly reduces computational overhead while maintaining expressiveness.

These codebook tokens function as \textbf{learnable placeholders} that the language model "writes into" during generation. As shown in Figure~\ref{fig:model_architecture}, they are integrated with visual features $\mathbf{f}_{\text{visual}}$ and trajectory representations $\mathbf{f}_{\text{Traj}}$ from the Trajectory Visual Perceiver to provide trajectory-guided spatial priors for segmentation.

\textbf{Multi-scale design rationale:} The 2-scale organization (3 tokens per scale) allows the model to capture both coarse object boundaries and fine-grained details:
\begin{itemize}
    \item \textbf{Scale 1 (coarse)}: 3 tokens encode global object shape and rough spatial layout
    \item \textbf{Scale 2 (fine)}: 3 tokens encode detailed boundaries and local geometric features
\end{itemize}

When the language model generates a \texttt{[SEG]} token during autoregressive decoding, we extract its corresponding embedding as trajectory-conditioned segmentation tokens $\mathbf{E}_{\text{seg}}$. This mechanism naturally interleaves segmentation with language generation: for example, when generating "The \texttt{[SEG]} object is on the table", the model produces text tokens and segmentation tokens in a unified sequence.

\subsection{Lightweight Pixel Decoder Architecture}

The decoder generates pixel-level masks as:
\begin{equation}
\hat{\mathbf{M}} = \mathcal{D}(\mathbf{f}_{\text{visual}}, \mathbf{E}_{\text{seg}})
\end{equation}
where $\hat{\mathbf{M}}$ denotes the predicted mask and $\mathcal{D}$ is a lightweight pixel decoder (shown as "Mask Decoder" in Figure~\ref{fig:model_architecture}).

Our decoder follows a simple yet effective 2-layer transformer architecture. This design contains only \textbf{12M parameters}, dramatically lighter than specialized segmentation decoders like SAM (636M parameters) or Mask2Former (223M parameters), while achieving competitive performance through trajectory-guided conditioning.

\subsection{Training Objectives and Loss Functions}

The model is trained end-to-end using an overall objective that combines three loss components:

\begin{equation}
\mathcal{L} = \mathcal{L}_{\text{txt}} + \lambda_{\text{ref}} \mathcal{L}_{\text{ref}} + \lambda_{\text{dice}} \mathcal{L}_{\text{dice}}
\end{equation}

where:
where:
\begin{itemize}
    \item $\mathcal{L}_{\text{txt}}$: Autoregressive cross-entropy loss for text generation (including segmentation codebook tokens). This ensures the model learns to generate \texttt{[SEG]} tokens at appropriate positions in the output sequence.
    
    \item $\mathcal{L}_{\text{dice}}$: DICE loss for mask overlap, computed as:
    \begin{equation}
    \mathcal{L}_{\text{dice}} = 1 - \frac{2|\hat{\mathbf{M}} \cap \mathbf{M}_{\text{gt}}|}{|\hat{\mathbf{M}}| + |\mathbf{M}_{\text{gt}}|}
    \end{equation}
    This loss is robust to class imbalance and encourages high mask IoU.
    
    \item $\mathcal{L}_{\text{ref}}$: Boundary refinement loss for accurate edge localization in overlapping regions. Denoting the mask predictions as $\{\hat{\mathbf{M}}_k \in \mathbb{R}^{H \times W}\}_{k=1}^{K}$ where $K$ is the total number of targets, $H$ and $W$ are the mask dimensions, and $\hat{M}_{k,i} \in \mathbb{R}$ represents the binary prediction at pixel $i$. We define a weight map $\mathbf{A}$ to assign increased attention to overlapping regions:
    \begin{equation}
    A_i = \begin{cases}
    \alpha, & \text{if } \sum_{k} \hat{M}_{k,i} \geq 2 \\
    1, & \text{if } \sum_{k} \hat{M}_{k,i} < 2
    \end{cases}
    \end{equation}
    where $\alpha$ is a hyper-parameter (set to 2.0 in our experiments). The weighted refinement loss is computed against the ground-truth mask $\mathbf{M}_k$ as:
    \begin{equation}
    \mathcal{L}_{\text{ref}} = \frac{1}{KHW} \sum_{k} \sum_{i} A_i \cdot \mathcal{L}_{\text{BCE}}(\hat{M}_{k,i}, M_{k,i})
    \end{equation}
    where $\mathcal{L}_{\text{BCE}}$ is per-pixel binary cross-entropy loss. This formulation emphasizes precise boundary delineation in challenging overlapping scenarios where multiple objects may compete for the same spatial region.
\end{itemize}

The loss coefficients are set to $\lambda_{\text{ref}} = 1.0$ and $\lambda_{\text{dice}} = 2.0$ to balance text generation quality with segmentation accuracy.

\textbf{Training resolution:} Masks are generated at 256×256 resolution during training and upsampled to original image resolution during inference. This resolution choice balances computational efficiency with segmentation quality.

\subsection{Comparison with Specialized Segmentation Decoders}

Table~\ref{tab:decoder_comparison} compares our lightweight decoder with specialized segmentation architectures. Despite having 53× fewer parameters than SAM and 19× fewer than Mask2Former, our decoder achieves competitive IoU performance while offering significant computational advantages.

\begin{table}[h]
\centering
\caption{Comparison of segmentation decoder architectures. Our lightweight 2-layer decoder achieves competitive performance with dramatically reduced parameters and latency.}
\label{tab:decoder_comparison}
\begin{tabular}{lccc}
\toprule
\textbf{Decoder Type} & \textbf{Parameters} & \textbf{RefCOCO IoU} & \textbf{Latency (ms)} \\
\midrule
SAM decoder & 636M & 84.2 & 245 \\
Mask2Former & 223M & 82.7 & 198 \\
\textbf{Ours (2-layer conv)} & \textbf{12M} & \textbf{83.4} & \textbf{107} \\
\bottomrule
\end{tabular}
\end{table}

Our approach achieves 83.4 IoU with only 12M parameters, approaching SAM's 84.2 IoU (which uses 636M parameters). The 56
\begin{enumerate}
    \item \textbf{Trajectory-guided priors}: As shown in Figure~\ref{fig:model_architecture}, trajectory information from the TVP module provides strong spatial cues that reduce the decoder's burden
    \item \textbf{Powerful backbone}: The Qwen2.5-VL backbone provides rich visual features that compensate for decoder simplicity
    \item \textbf{End-to-end training}: Joint optimization allows the decoder to specialize for trajectory-conditioned segmentation
\end{enumerate}

\subsection{Codebook Size Ablation Study}

Table~\ref{tab:codebook_ablation} analyzes the trade-off between codebook size and segmentation quality. We experiment with 3, 6, and 12 learnable tokens.

\begin{table}[h]
\centering
\caption{Ablation study on segmentation codebook size. 6 tokens provide optimal balance between performance and parameter efficiency.}
\label{tab:codebook_ablation}
\begin{tabular}{ccc}
\toprule
\textbf{\# Tokens} & \textbf{RefCOCO IoU} & \textbf{Parameters} \\
\midrule
3 & 80.8 & 6M \\
6 (ours) & 83.4 & 12M \\
12 & 83.9 & 24M \\
\bottomrule
\end{tabular}
\end{table}

We choose 6 tokens as our default configuration because:
\begin{itemize}
    \item \textbf{Diminishing returns}: Increasing from 6 to 12 tokens yields only +0.5 IoU improvement while doubling parameters (12M → 24M)
    \item \textbf{Performance sufficiency}: 83.4 IoU is competitive with specialized systems and sufficient for most referring segmentation tasks
    \item \textbf{Computational efficiency}: 12M parameters maintain fast inference while preserving quality
\end{itemize}

The 3-token configuration shows significant performance degradation (-2.6 IoU), suggesting that a minimum level of representational capacity is necessary to capture diverse object shapes and boundaries.

\subsection{Extension to Video Segmentation}

Our segmentation design naturally generalizes to video segmentation by applying the decoder on multi-frame features $\{\mathbf{f}^t_{\text{visual}}\}_{t=1}^{T}$ with trajectory guidance across frames. The temporal consistency of human gaze trajectories provides strong priors for tracking objects across video frames:

\begin{equation}
\hat{\mathbf{M}}_t = \mathcal{D}(\mathbf{f}^t_{\text{visual}}, \mathbf{E}^t_{\text{seg}}), \quad t = 1, \ldots, T
\end{equation}

where trajectory representations naturally encode temporal relationships between frames. This enables TraceVision to perform video referring segmentation without additional architectural modifications or temporal modeling components.

\section{System Prompt for Baseline Models}

To enable baseline LVLMs to understand and utilize trajectory information, we provide comprehensive system prompts for different task scenarios. Each prompt is carefully designed to match the actual trajectory format used in our dataset.

\begin{tcolorbox}[colback=blue!5, colframe=blue!50!black, title=\textbf{System Prompt: General Trajectory Understanding}]
\small
You are a vision-language model with trajectory understanding capability.

\textbf{Trajectory Format:} 
\begin{itemize}
\item \textbf{Input format:} \texttt{<Trajectory>(x1,y1), (x2,y2), (x3,y3), ... </Trajectory>}
\item \textbf{Output format:} \texttt{<Traj>[x1,y1]</Traj> <Traj>[x2,y2]</Traj> ...}
\end{itemize}

\textbf{Format Specification:}
\begin{itemize}
\item \texttt{(xi, yi)} or \texttt{[xi, yi]}: Pixel coordinates in the image
\item Points represent sequential spatial locations in the attention path
\item The order of points indicates the temporal sequence of visual exploration
\end{itemize}

\textbf{Semantic Interpretation:}\\
The trajectory represents sequential human attention movement across the image. Dense clusters of points indicate regions of high interest with prolonged attention, while sparse regions indicate rapid scanning. The point sequence reflects the natural flow of visual exploration.

\textbf{Guidelines:}
\begin{itemize}
\item Use trajectory information to identify focus regions
\item Pay attention to the sequential order of points
\item Consider spatial distribution patterns to determine regions of interest
\item Integrate trajectory information with visual features for accurate responses
\end{itemize}
\end{tcolorbox}

\vspace{0.5cm}

\begin{tcolorbox}[colback=green!5, colframe=green!50!black, title=\textbf{Task Prompt 1: Trajectory-Guided Captioning (Referential Trajectory Interpretation)}]
\small
\textbf{Task Description:}\\
Given an image and a trajectory representing human attention movement, generate a natural language description of what the trajectory focuses on.

\textbf{Input Format:}\\
\texttt{Image: [IMAGE]}\\
\texttt{Instruction: User's gaze followed this trajectory across the image:}\\
\texttt{<Trajectory>(x1,y1), (x2,y2), ..., (xN,yN) </Trajectory>}

\textbf{Task Requirements:}
\begin{itemize}
\item Focus on objects and regions indicated by the trajectory path
\item Include object identity, attributes, and spatial relationships
\item Generate coherent descriptions following the attention flow
\item Describe the overall scene and specific elements along the trajectory
\item Do not mention the trajectory format in your response
\end{itemize}

\textbf{Output Format:}\\
Natural language caption describing what the trajectory focuses on.

\textbf{Example 1:}\\
\textit{Input:}\\
\texttt{User's gaze followed this trajectory across the image:}\\
\texttt{<Trajectory>(350,150), (360,300), (410,210), (510,350), (50,380), (600,380), (10,50), (600,200) </Trajectory>}

\textit{Output:}\\
"In the center of the image there is a man standing and holding a ball in his hand. On the right there is another man bending. In the background there is a fence, trees and building."

\textbf{Example 2:}\\
\textit{Input:}\\
\texttt{User's gaze followed this trajectory across the image:}\\
\texttt{<Trajectory>(325,230), (293,185), (284,170), (220,75), (325,230) </Trajectory>}

\textit{Output:}\\
"Behind the boy wearing red shoes, two boys dressed in white tops and gray pants are standing together on the grass, looking into the distance. On the right, a boy wearing a white jersey with number 16 and holding a frisbee, along with another boy in a red jersey with number 4, are running back and forth."
\end{tcolorbox}

\vspace{0.5cm}

\begin{tcolorbox}[colback=orange!5, colframe=orange!50!black, title=\textbf{Task Prompt 2: Caption-Guided Trajectory Generation (Referential Trajectory Grounding)}]
\small
\textbf{Task Description:}\\
Given an image and a natural language caption describing specific content, predict the human attention trajectory that would naturally occur when observing the described content.

\textbf{Input Format:}\\
\texttt{Image: [IMAGE]}\\
\texttt{Instruction: If you were to describe, "[caption content]," please generate a possible gaze trajectory.}

\textbf{Task Requirements:}
\begin{itemize}
\item Generate trajectory coordinates that align with the described content's spatial location
\item Output each trajectory point in the format \texttt{<Traj>[x,y]</Traj>}
\item Create sequential points that follow a natural viewing path
\item Include multiple points to represent the attention flow across described objects
\item Ensure coordinates correspond to actual locations of mentioned objects
\end{itemize}

\textbf{Output Format:}\\
\texttt{<Traj>[x1,y1]</Traj> <Traj>[x2,y2]</Traj> ... <Traj>[xN,yN]</Traj>}

\textbf{Example 1:}\\
\textit{Input:}\\
\texttt{If you were to describe, "In this image there is one plate and on that plate there is one tissue paper and one Sandwich and one lemon slice is there," please generate a possible gaze trajectory.}

\textit{Output:}\\
\texttt{<Traj>[80,200]</Traj> <Traj>[100,180]</Traj> <Traj>[130,170]</Traj> <Traj>[170,160]</Traj> <Traj>[210,130]</Traj>}

\textbf{Example 2:}\\
\textit{Input:}\\
\texttt{If you were to describe, "City street with neatly lined-up shops, cars parked along the roadside, restaurant signs and logos, customers sitting in front of storefronts, and an airplane flying in the sky," please generate a possible gaze trajectory.}

\textit{Output:}\\
\texttt{<Traj>[80,200]</Traj> <Traj>[100,180]</Traj> <Traj>[130,170]</Traj> <Traj>[170,160]</Traj> <Traj>[210,130]</Traj> <Traj>[300,60]</Traj>}
\end{tcolorbox}

\vspace{0.5cm}

\begin{tcolorbox}[colback=purple!5, colframe=purple!50!black, title=\textbf{Task Prompt 3: Trajectory-Based Visual Question Answering (Interactive Trajectory Reasoning QA)}]
\small
\textbf{Task Description:}\\
Given an image, one or more trajectories representing human attention, and a question, answer the question based on the regions indicated by the trajectory.

\textbf{Input Format:}\\
\texttt{Image: [IMAGE]}\\
\texttt{Trajectory: <Trajectory>(x1,y1), (x2,y2), ..., (xN,yN) </Trajectory>}\\
\texttt{Question: [Question about trajectory-indicated content]}

\textbf{Task Requirements:}
\begin{itemize}
\item Analyze the trajectory to identify the regions being focused on
\item Extract relevant visual information from trajectory-indicated areas
\item Provide accurate and concise answers
\item When comparing multiple trajectories, identify which best matches the question
\item Use spatial patterns and point density to determine focus regions
\end{itemize}

\textbf{Output Format:}\\
Concise answer based on trajectory-indicated regions.

\textbf{Example 1:}\\
\textit{Input:}\\
\texttt{Given Trajectory A <Trajectory>(350,150), (360,300), (410,210)</Trajectory> and Trajectory B <Trajectory>(50,380), (600,380), (10,50), (600,200)</Trajectory>, which one is more likely to represent a person describing the baseball in this image?}

\textit{Output:}\\
"The answer is Trajectory A"

\textbf{Example 2:}\\
\textit{Input:}\\
\texttt{User's gaze followed this trajectory across the image: <Trajectory>(350,150), (360,300), (410,210), (510,350) </Trajectory>. Based on this path, what objects are being focused on in the scene?}

\textit{Output:}\\
"The trajectory focuses on a red car parked on the street and a person standing nearby."

\textbf{Example 3:}\\
\textit{Input:}\\
\texttt{When seeing a house in the distant background with a tree next to it and a bench beside the tree, how should the human attention trajectory be located?}

\textit{Output:}\\
\texttt{<Trajectory>(112,345), (118,342), (123,310), (125,300), (127,290), (140,275), (142,280), (145,285), (148,290)</Trajectory>}

\textbf{Example 4:}\\
\textit{Input:}\\
\texttt{Which of the following trajectories best corresponds to the description: "athletes competing for the disc"?}\\
\texttt{A: <Trajectory>(x1,y1), (x2,y2), ...</Trajectory>}\\
\texttt{B: <Trajectory>(x1,y1), (x2,y2), ...</Trajectory>}

\textit{Output:}\\
"The correct answer is Trajectory B."
\end{tcolorbox}

\vspace{0.3cm}

\begin{figure*}[!ht]
    \centering
    \includegraphics[width=1\textwidth]{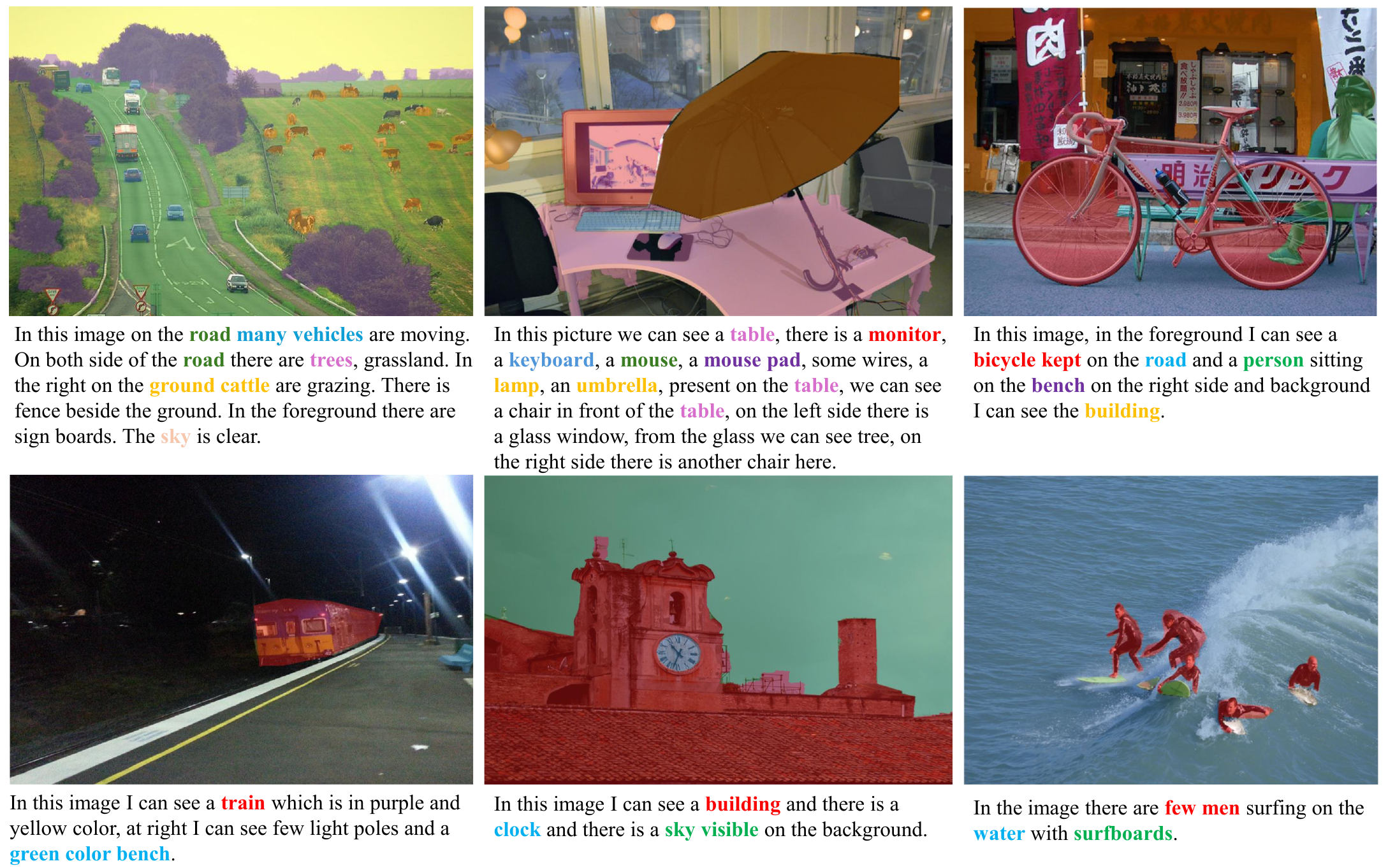}
    \caption{\textbf{Qualitative segmentation and description results.} 
    Examples showing color-coded segmentation masks and corresponding natural language descriptions across diverse scenarios including outdoor scenes, indoor workspaces, urban settings, transportation, architecture, and action scenes. The model accurately identifies objects (highlighted in different colors) and generates detailed descriptions with correct spatial relationships and object attributes.}
    \label{fig:segmentation_results}
\end{figure*}

\begin{figure*}[!ht]
    \centering
    \includegraphics[width=1.0\textwidth]{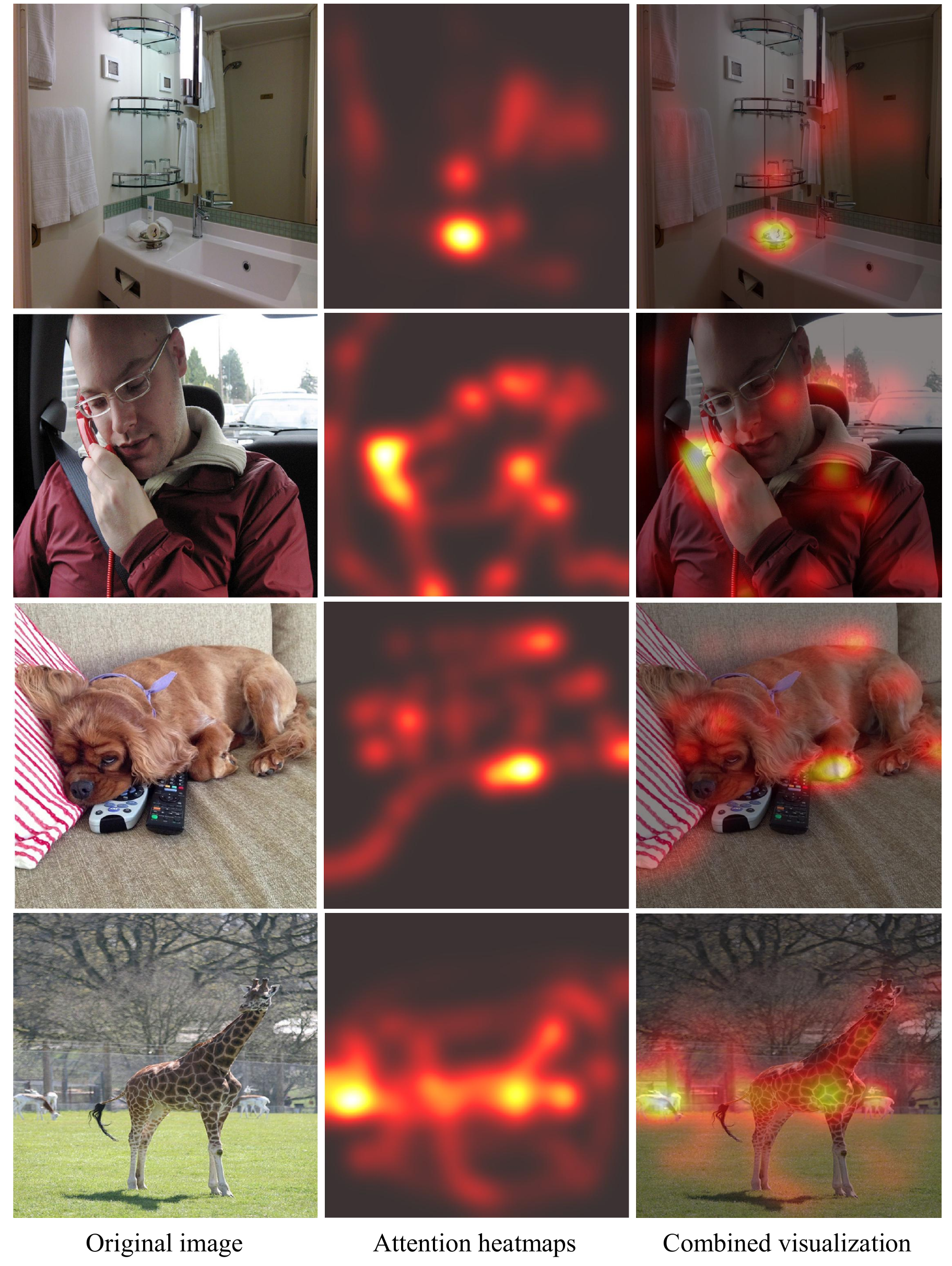}
    \caption{\textbf{Attention heatmap visualization.} 
    Visualization of the Trajectory Visual Perceiver's attention mechanism across four examples. Each row shows the original image (left), attention heatmap with red-to-yellow gradient indicating attention intensity (middle), and overlay visualization (right). The heatmaps reveal that the model focuses on semantically salient objects and regions relevant for scene understanding.}
    \label{fig:attention_heatmaps}
\end{figure*}

\section{Additional Qualitative Results}
\label{app:qualitative}

\subsection{Segmentation and Description Examples}
\label{app:segmentation}

Figure~\ref{fig:segmentation_results} demonstrates TraceVision's joint visual-linguistic understanding across six diverse scenarios. The model successfully performs object segmentation with color-coded masks while generating accurate natural language descriptions. Key observations include:

\textbf{Multi-scale object detection:} The model handles objects of varying sizes, from large structures (buildings, trains) to small items (mice, keyboards, remote controls).

\textbf{Spatial reasoning:} Generated descriptions accurately capture spatial relationships using phrases like "on both side of the road," "in the foreground," "at right," and "on the bench on the right side."

\textbf{Attribute recognition:} The model correctly identifies object attributes including colors ("purple and yellow color" train, "green color bench"), states ("cattle are grazing," "kept on the road"), and quantities ("many vehicles," "few men").

\textbf{Scene diversity:} Examples span outdoor environments (road scenes with vehicles and livestock), indoor settings (office workspace), urban contexts (bicycles and buildings), transportation infrastructure, architectural landmarks, and dynamic action scenes.

\subsection{Attention Mechanism Analysis}
\label{app:attention}

Figure~\ref{fig:attention_heatmaps} visualizes the internal attention patterns of the Trajectory Visual Perceiver module. The attention heatmaps reveal several important characteristics:

\textbf{Selective focus on salient objects:} The model concentrates attention on semantically meaningful objects (sink and shelves in bathroom, person's face and phone in vehicle, dog's body and remote controls, giraffe's body) rather than background regions.

\textbf{Multi-region attention:} In complex scenes, attention is distributed across multiple relevant regions (e.g., person's face and hand holding phone; giraffe's body and ground plane), indicating the model's ability to capture relationships between objects.

\textbf{Context-aware activation:} Attention patterns adapt to scene context—functional elements receive higher attention in indoor scenes, while primary subjects dominate in portrait-style images and outdoor wildlife scenes.

These visualizations demonstrate that the TVP effectively learns to attend to regions critical for trajectory prediction and scene understanding tasks.













\end{document}